%% file: paper.tex
\pdfoutput=1

\documentclass[11pt]{article}
\usepackage{setspace}
\usepackage[hidelinks]{hyperref}

\input{math.tex}

\usepackage{style}

\usepackage[]{acl}

\usepackage{times}
\usepackage{latexsym}
\usepackage{textcomp}
\usepackage{cuted}

\usepackage[T1]{fontenc}

\usepackage[utf8]{inputenc}

\usepackage{microtype}

\usepackage{inconsolata}

\usepackage[]{mdframed}

\newcommand{\bok}{best-of-$n$\xspace}

\definecolor{orange}{rgb}{1,0.5,0}
\definecolor{mdred}{rgb}{0.7,0,0}
\definecolor{mdgreen}{rgb}{0.05,0.6,0.05}
\definecolor{mdblue}{rgb}{0,0,0.7}
\definecolor{dkblue}{rgb}{0,0,0.5}
\definecolor{dkgreen}{rgb}{0,0.5,0}
\definecolor{dkgray}{rgb}{0.3,0.3,0.3}
\definecolor{slate}{rgb}{0.25,0.25,0.4}
\definecolor{gray}{rgb}{0.5,0.5,0.5}
\definecolor{ltgray}{rgb}{0.7,0.7,0.7}
\definecolor{purple}{rgb}{0.7,0,1.0}
\definecolor{lavender}{rgb}{0.65,0.55,1.0}
\definecolor{theme}{HTML}{6b8a97}

\newcommand{\papercomment}[3]{\ensuretext{\textcolor{#3}{[#1 #2]}}}
\renewcommand{\papercomment}[3]{}  %

\newcommand{\zerodisplayskips}{%
  \setlength{\abovedisplayskip}{7pt}%
  \setlength{\belowdisplayskip}{7pt}%
  \setlength{\abovedisplayshortskip}{7pt}%
  \setlength{\belowdisplayshortskip}{7pt}}
\appto{\normalsize}{\zerodisplayskips}
\appto{\small}{\zerodisplayskips}
\appto{\footnotesize}{\zerodisplayskips}

\newif\iflongversion
\longversiontrue

\title{Reuse Your Rewards:\\Reward Model Transfer for Zero-Shot Cross-Lingual Alignment}

\author{Zhaofeng Wu$^\text{\Cancer}$ \hskip0.4em\relax
    Ananth Balashankar$^\text{\Virgo}$ \hskip0.4em\relax
    Yoon Kim$^\text{\Cancer}$ \hskip0.4em\relax
    Jacob Eisenstein$^\text{\Virgo}$ \hskip0.4em\relax
    Ahmad Beirami$^\text{\Virgo}$\\
    $^\text{\Cancer}$MIT \quad $^\text{\Virgo}$Google DeepMind \\
    \texttt{zfw@csail.mit.edu}
}

\begin{document}
\maketitle

\begin{abstract}
Aligning language models (LMs) based on human-annotated preference data is a crucial step in obtaining practical and performant LM-based systems.
However, multilingual human preference data are difficult to obtain at scale, making it challenging to extend this framework to diverse languages.
In this work, we evaluate a simple approach for zero-shot cross-lingual alignment, where a reward model is trained on preference data in one source language and directly applied to other target languages.
On summarization and open-ended dialog generation, we show that this method is consistently successful under comprehensive evaluation settings, including human evaluation: cross-lingually aligned models are preferred by humans over unaligned models on up to >70\% of evaluation instances.
We moreover find that a different-language reward model sometimes yields better aligned models than a same-language reward model.
We also identify best practices when there is no language-specific data for even supervised finetuning, another component in alignment.
\iflongversion
\blfootnote{Work done while ZW was a part-time intern at Google.}
\fi
\end{abstract}

\input{sections/01_intro}

\input{sections/02_background}
\input{sections/03_method}

\input{sections/04_setup}
\input{sections/05_results}
\input{sections/06_analysis}

\input{sections/07_translated_sft}
\input{sections/08_related_work}
\input{sections/09_conclusion}

\section*{Limitations}

Free-form generation is challenging to evaluate, especially in a cross-lingual setup. As we mentioned, neither the finetuned target-language RM evaluator scores nor pairwise evaluation from humans or LMs are perfect~\citepia{wang2023how,zheng2023large,hosking2024human}.
Nevertheless, we believe the consistent cross-lingual transferability observed across our many evaluation settings suggests that it would hold more generally. Similarly, it is not possible to comprehensively study the myriad of reward optimization methods~\citepia{rafailov2023direct,azar2023general}, some of which may not enjoy the same cross-lingual RM transfer benefit (in fact, the notion of a RM do not even exist in some, though analogous ideas may be applicable).
However, the two that we study, \bok and PPO, are representative of current common practices, especially given the strong empirical performance of \bok~\citepia{gao2022scaling,mudgal2023controlled,rafailov2023direct}.
Somewhat orthogonally, past work has argued that it is limiting to use one single scalar to represent generation quality~\citep{xu-etal-2023-critical,krishna-etal-2023-longeval,hosking2024human} and that more fine-grained rewards could be beneficial~\citep{wu2023finegrained}. We follow the convention to use one single score to more easily measure and compare cross-lingual transfer in many setups, but a similar but more fine-grained study would be valuable future work.
It has also been shown that it is more challenging to train reward models for low-resourced languages~\citep{shen2024language}.
We only considered relatively high-resourced languages in this work, and it is possible that the pattern would differ when using lower-resourced source languages for transfer.
Finally, our motivating assumption that generation quality being language-agnostic does not always hold, especially when facing culture-specific tasks or task instances. In those cases, we believe we would see reduced cross-lingual generalizability.

\iflongversion
\section*{Acknowledgments}
We would like to thank Jonathan Berant, Jilin Chen, Elizabeth Clark, Daphne Domansi, Jie Fan, Han Guo, Henry Hand, Harrison Lee, Jong Lee, Alisa Liu, Ana Marasović, Usha Rani Markuk, Joshua Maynez, Kathy Meier-Hellstern, Chirag Nagpal, Flavien Prost, Linlu Qiu, Kevin Robinson, Alexis Ross, Shannon Zejiang Shen, Bailin Wang, Xinyan Velocity Yu, and the T5X team Google for their valuable feedback and support. The MIT researchers were partially supported by funds from an MIT-IBM Watson AI Lab grant.
\fi

\bibliography{custom}

\clearpage
\appendix

\input{sections/99_appendix}

\end{document}

%% file: math.tex
\usepackage{amsmath,amsfonts,amsthm,amssymb,bm}

\def\eqref#1{equation~\ref{#1}}

\def\1{\bm{1}}

\DeclareMathAlphabet{\mathsfit}{\encodingdefault}{\sfdefault}{m}{sl}
\SetMathAlphabet{\mathsfit}{bold}{\encodingdefault}{\sfdefault}{bx}{n}

\newcommand{\E}{\mathbb{E}}

\theoremstyle{definition}

\theoremstyle{definition}

%% file: sections/01_intro.tex
\section{Introduction}

Alignment has become an indispensable stage for building practical language models (LMs) adjusted to human preferences.
This additional step, however, makes it challenging to develop LMs for many languages:
unlike for autoregressive language modeling where multilingual unlabeled data may be easy to obtain~\citep{joshi-etal-2020-state}, such as religious texts~\citep{Christodouloupoulos2015}, labeled preference data can be expensive to gather. 
How do we align a LM in a target language without any preference data in that language?

\begin{figure}[!hp]
    \vspace{-0.2cm}
    \centering
    \includegraphics[width=0.48\textwidth]{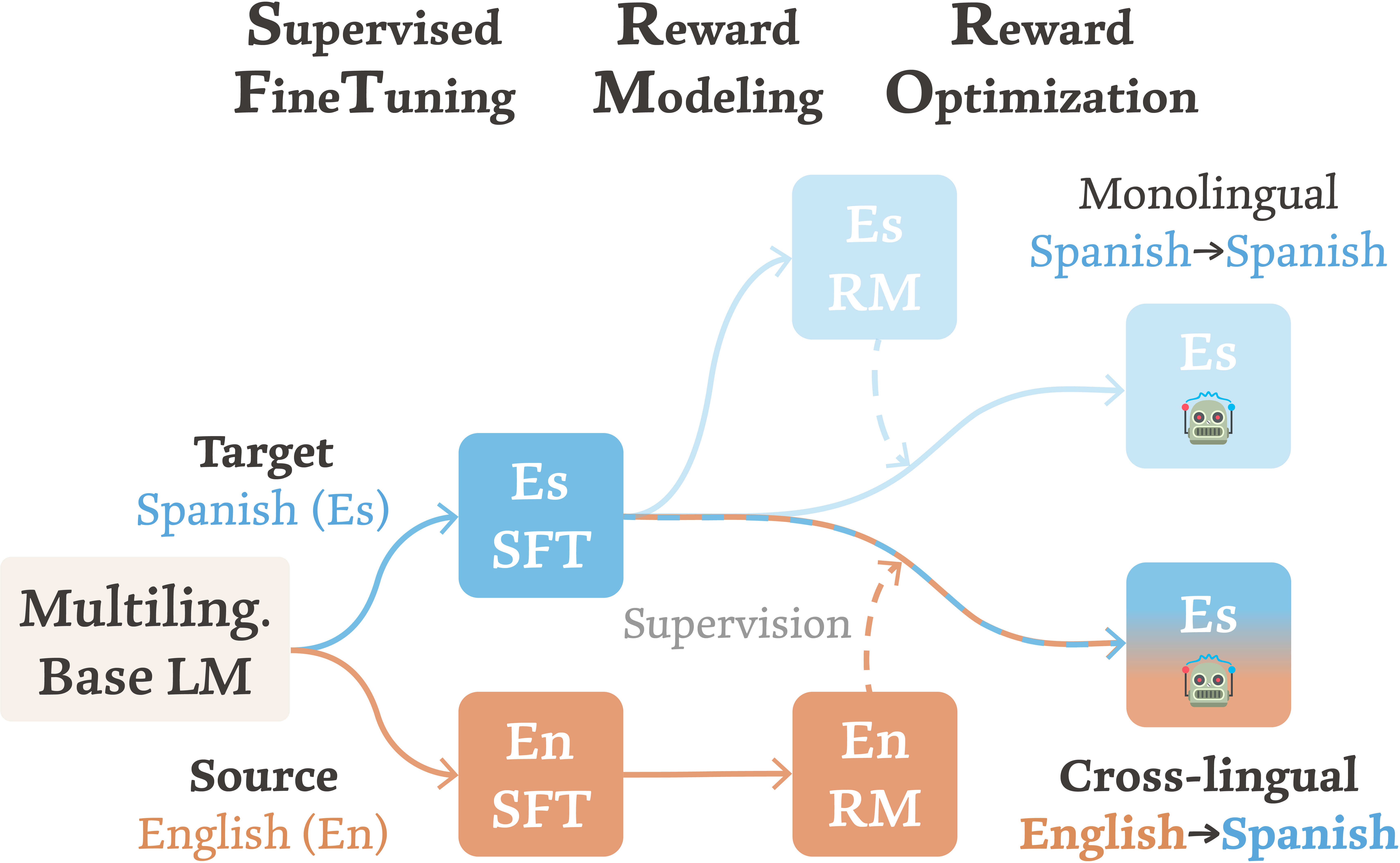}
    \caption{Cross-lingual reward model (RM) transfer. To align in a target language (in this example, Spanish), common monolingual alignment uses a RM for that target language. Instead, we re-purpose a RM for a different source language (in this example, English).}
    \label{fig:overview}
\end{figure}

\begin{figure}[!hp]
    \vspace{-0.2cm}
    \centering
    \includegraphics[width=0.48\textwidth]{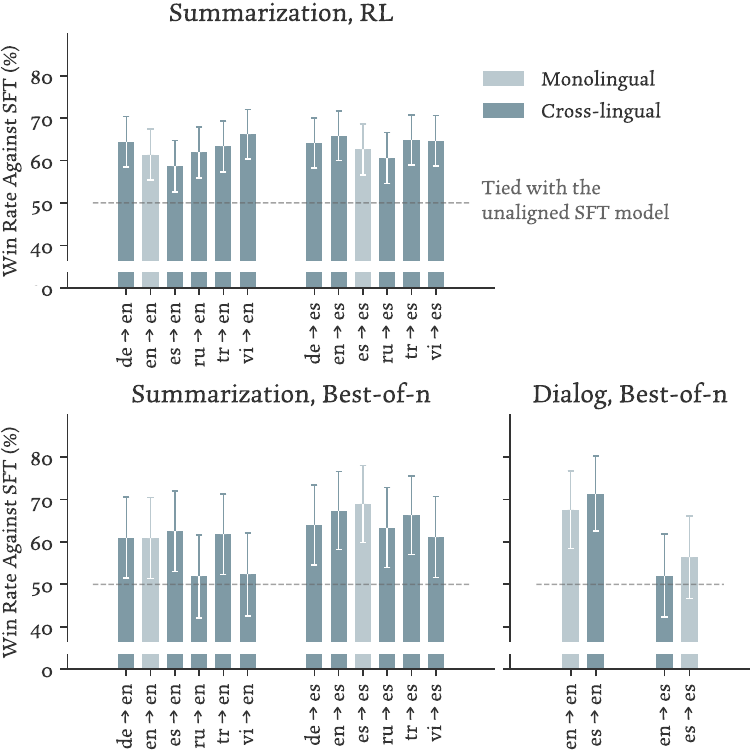}
    \captionof{figure}{\textbf{Performing target-language alignment using a RM for a different source language improves performance, when evaluated exclusively in the target language.
    This improvement is sometimes even larger than using the target-language RM} (monolingual alignment).
    Here we measure the win rate against the target-language (unaligned) SFT model judged by humans, and the 95\% confidence interval across validation instances. ``source$\to$target`` denotes using a source-language RM to drive alignment in the target language.
    }
    \vspace{-0.2cm}
    \label{fig:humans}
\end{figure}

We propose a novel reward model (RM) transfer setup, where we re-purpose a trained RM for some source language to align a LM in a target language (\autoref{fig:overview}), and investigate the effectiveness of this simple recipe.
Across two tasks (summarization and open-ended dialog generation), two reward optimization methods (reinforcement learning and \bok reranking), and various evaluation settings, we demonstrate substantial and consistent zero-shot cross-lingual utility of RMs.
Surprisingly, alignment using a different-language RM sometimes outperforms using a same-language RM, both when judged by humans and LMs.
We also show that our RM transfer framework is useful even when target-language data for supervised finetuning (SFT), another component in alignment, is inaccessible.

Our results show that RM signals are generalizable and robust to input distribution changes, which could be leveraged for more future applications.
Practically, our findings pave the path towards lowering the costs for training and deploying LMs that more equitably serve users around the world.

%% file: sections/02_background.tex
\section{Background\iflongversion: Alignment From Human Feedback\fi} \label{sec:background}

In addition to traditional unsupervised LM pretraining, many recent LMs also include an alignment phase to improve helpfulness, harmlessness, etc., supervised by human feedback~\citepia{bai2022training,ouyang2022training}.
A common recipe includes three stages: supervised finetuning (SFT), reward modeling (RM), and reward optimization. We give an overview of each and refer readers to \citet{ouyang2022training} and \citet{bai2022training} for details. We assume a \textbf{base} model already pretrained using a usually next-token prediction objective.

\paragraph{The SFT stage} initializes from the base model and takes task inputs $x\in\mathcal{X}$ to train the model to simulate example outputs $y\in\mathcal{Y}$. Specifically, it optimizes the conditional log-likelihood of $y$ given some input $x$, similar to regular language modeling. We denote the trained SFT model using $\pi_{\text{SFT}}$.

\paragraph{The RM stage} trains a model ${r: \mathcal{X} \times \mathcal{Y} \to \mathbb{R}}$ as a proxy for human-judged quality of $y$ under $x$.
It initializes from $\pi_{\text{SFT}}$ and is trained using a dataset of human judgments of generations.
We consider two types of feedback to train the RM:

\begin{enumerate}[leftmargin=*,itemindent=0pt]
    \item \textbf{Pointwise} feedback judges the quality of a single generation; in particular we only consider binary (good or bad) pointwise judgments. Denoting it as $z\in\{0,1\}$ and letting $\mathcal{D}^{\text{RM}}$ be a dataset of judgments, the RM can be a standard classifier trained using the cross-entropy loss,
    \begin{align*}
    \hspace{-.2in}-\E_{(x,y,z)\sim\mathcal{D}^{\text{RM}}}[&z \log \sigma \left(r(x,y)\right) + \\
    &(1 - z) \log \left(1- \sigma \left(1 - r(x,y)\right)\right)].
    \end{align*}
    
    \item \textbf{Pairwise} feedback chooses a better generation out of two. We denote the chosen one as $y_w$ and the other as $y_l$. To train a pointwise RM on such data, the Bradley-Terry model~\citep{bradley-terry} is often used, maximizing
    \begin{align*}
    \E_{(x,y_w,y_l)\sim\mathcal{D}^{\text{RM}}}[\log \sigma \left(r(x, y_w) - r(x, y_l)\right)].
    \end{align*}
    It is also generalizable to more than two outputs.
\end{enumerate}

\paragraph{The reward optimization stage} also initializes from $\pi_{\text{SFT}}$ and further adjusts the model outputs using human feedback (as captured by the RM). Two common methods are reinforcement learning (RL) and \bok.
Best-of-$n$  is an inference-time procedure that does not change the underlying model, where multiple generations are sampled from $\pi_{\text{SFT}}$ and then reranked using the RM; the highest-scoring generation is returned as the output.
In RL, the model itself is changed such that its samples are scored highly by the RM, with the objective
\hspace{-.2in}
\resizebox{\linewidth}{!}{
\begin{minipage}{\linewidth}
  \vspace{-0.1in}
  \begin{align*}
\E_{x\sim\mathcal{D}^{\text{RO}},\tilde{y}\sim\pi_{\theta}(x)}[&r(x, \tilde{y}) - \\
&\beta \left(\log \pi_\theta (\tilde{y}\mid x) - \log \pi_{\text{SFT}} (\tilde{y} \mid x)\right)].
\end{align*}
\vspace{-0.15in}
  \end{minipage}
}
$\mathcal{D}^{\text{RO}}$ is a dataset of inputs and $\beta$ is a regularization hyperparameter. The above is typically optimized with  PPO~\citep{schulman2017proximal}.
While we generally experiment with both methods, in some of our analyses we focus on \bok for a clean testbed without confounders from RL training.

%% file: sections/03_method.tex
\section{Reward Model Transfer for Cross-Lingual Alignment} \label{sec:method}

The pipeline in \S\ref{sec:background} is usually performed monolingually, commonly in English. Aligning for a new language requires both SFT data and RM data in that language. While the former may be relatively easier to obtain due to automatic construction methods, such as by re-purposing existing multilingual datasets~\citep{muennighoff-etal-2023-crosslingual} or by eliciting from LMs~\citep{wang-etal-2023-self-instruct}, RM data for a new language can be more expensive to gather, as it in principle requires human judgments. Additionally, RM data should ideally be periodically re-collected to avoid over-optimization~\citep{bai2022training}, further increasing data demand.
Thus, we are mainly interested in alignment without target-language RM data, though, in \S\ref{sec:translated-sft}, we investigate dispensing with target-language SFT data too.

We propose to perform reward optimization using a RM trained for a different language~(\autoref{fig:overview}).
Intuitively, assuming model generation quality transfers cross-lingually (e.g., good English generations are still good when translated into Spanish\footnote{We believe this is a weak assumption, though for tasks and instances more subject to culture-specific factors, generations may be judged more differently across languages~\citep{costa2014your,hershcovich-etal-2022-challenges,shwartz-2022-good}. 
}), a model that can judge the output quality in one language should generalize to others, as long as the RM understands the languages, which is enabled by multilingual \emph{base} model training.
This generalizability is often observed for other tasks in the zero-shot cross-lingual transfer literature~\citepia{wu-dredze-2019-beto,pires-etal-2019-multilingual,conneau-etal-2020-emerging,pmlr-v119-hu20b}, and we expect it to work for RMs too.
A simple baseline would be to use automatically translated RM data, to which we compare in \S\ref{sec:transferability-as-alignment-signal}.
In this paper, we use \textbf{source language} to denote the RM language, and \textbf{target language} for the language of the aligned model.

%% file: sections/04_setup.tex
\section{Experimental Setup} \label{sec:setup}

We consider two tasks: summarization, common in alignment research~\citepia{stiennon2020learning,ziegler2020finetuning,lee2023rlaif}, and open-ended dialog generation, with substantial real-world relevance.
\S\ref{sec:dataset-stats} describes dataset details and statistics. \S\ref{sec:training-details} includes training details.
\S\ref{sec:task-instructions} contains our task instructions.

\paragraph{Summarization.}
The Seahorse dataset~\citep{clark-etal-2023-seahorse} contains documents and summaries in six languages (German, English, Spanish, Russian, Turkish, and Vietnamese) with pointwise human ratings which we use.
For SFT, we gather the data sources of Seahorse: XSum~\citep{narayan-etal-2018-dont}, XL-Sum~\citep{hasan-etal-2021-xl}, MLSum~\citep{scialom-etal-2020-mlsum}, and WikiLingua~\citep{ladhak-etal-2020-wikilingua}.
We use mT5-XL~\citep{xue-etal-2021-mt5} as our multilingual base model, with 3.7B parameters.

\paragraph{Open-Ended Dialog Generation.}
We use the OpenAssistant dataset~\citep{kopf2023openassistant} with multilingual, pairwise human-rated chat transcripts.\footnote{In \url{https://huggingface.co/datasets/OpenAssistant/oasst1}.}
For the SFT data, we use the human-preferred response in each pair to finetune the model.
Many languages in OpenAssistant have only limited data, so we only consider three languages with the most amounts of data: English, Spanish, and Russian.
We use PaLM-2-XXS as the base model~\citep{anil2023palm}. The authors of OpenAssistant found RL to be ineffective for this dataset~\citep{kopf2023openassistant}, which we confirmed in our experiments~(\autoref{fig:palm}). We therefore focus on \bok for this task.

\paragraph{Evaluation.}
We assess model quality across several settings. 
First, we use the target-language RM, which is by design finetuned to judge target-language generation quality.
But because of potential RM biases~\citep{gao2022scaling,coste2023reward,eisenstein2023helping}, we also include two zero-shot-prompted evaluation models with much larger backbones---GPT-4~\citep{gpt4} and PaLM-2-L~\citep{anil2023palm}. This latter evaluation setup is common in prior work and has been demonstrated to correlate well with human judgments~\citepia{lee2023rlaif,rafailov2023direct,an2023leval,mu2023learning}. We also confirm its validity in \S\ref{sec:transferability-as-alignment-signal} and \S\ref{sec:lm-acc-on-rm-data}.
Importantly, both evaluation LMs support multilingual texts.
Finally, we also perform human evaluations by self-reported native or advanced speakers, though only for a subset of language pairs and 250 (RL) / 100 (\bok) instances per pair due to its cost.
For both human and LM evaluation, we elicit pairwise judgments to compare responses from the aligned model and the SFT model~\citepia{bai2022constitutional,lee2023rlaif}. We measure the \emph{win rate}, i.e., how often the judge prefers the former. A 50\% win rate indicates no improvement from alignment.
\S\ref{sec:lm-eval-prompts} includes more details such as the evaluation prompts and positional bias control.

%% file: sections/05_results.tex
\section{Results} \label{sec:results}

Here we report the results of cross-lingual alignment.
See \S\ref{sec:raw-results} for numerical results that correspond to the plots in this section.

\subsection{Cross-Lingual Alignment Is Effective} \label{sec:transferability-as-alignment-signal}

\begin{figure*}[t!]
    \centering
    \includegraphics[width=0.9\textwidth]{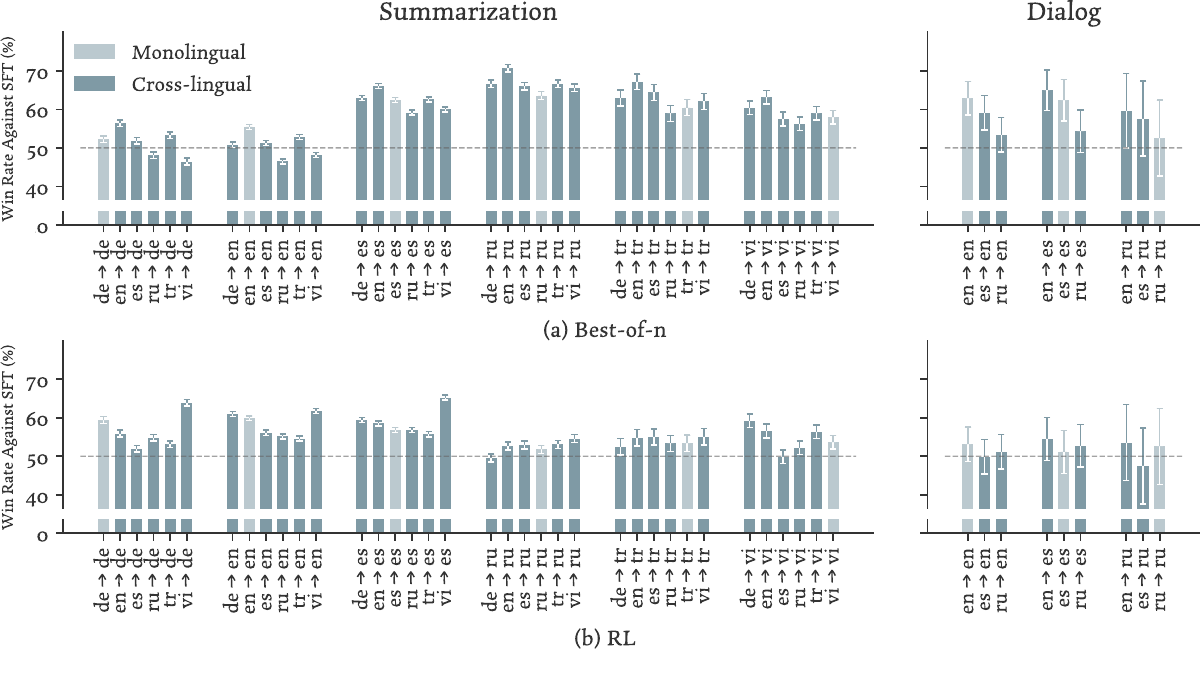}
    \vspace{-3mm}
    \caption{Alignment effectiveness, compared to the target-language SFT model judged by PaLM-2-L, and the 95\% confidence interval across validation instances. ``source$\to$target`` denotes a source-language RM driving alignment in the target language. \textbf{Cross-lingual alignment is generally effective, sometimes outperforming monolingual alignment}. RL is hard to train for OpenAssistant, in line with what its authors found~\citep{kopf2023openassistant}.}
    \vspace{-3mm}
    \label{fig:palm}
\end{figure*}

When evaluated by the finetuned target-language RM, \autoref{fig:training-dynamic-crosslingual} shows that monolingual \bok or RL always improves model quality, as expected.
Encouragingly, cross-lingual reward optimization improves over the SFT model in all cases too.
Similarly, when judged by a general-purpose LM, PaLM-2-L in \autoref{fig:palm} and GPT-4 in \S\ref{sec:gpt4-results},  in-language and cross-lingual reward optimization both generally improve model quality.
Importantly, we observe high agreement between the two LMs: on an instance level, they agree >70\% across setups (see \S\ref{sec:gpt4-results}); if we consider how often they agree in the relative ranking of two source languages, they agree 78\% for summarization (both \bok and RL) and 100\% for dialog generation (\bok). This indicates the reliability of a LM judge.

Human evaluation (\autoref{fig:humans}) reveals the same trend, though with larger confidence intervals due to the cost.
Human evaluation results also validate and justify LM-based evaluation: For summarization, PaLM-2-L (GPT-4) agrees with humans 65\% (69\%) of the time in English and 66\% (62\%) in Spanish, matching the 63\% human-human agreement for English reference summaries and 67\% for Spanish in Seahorse
\iflongversion(Clark, personal communication, April 15, 2024)\fi. For dialog, PaLM-2-L (GPT-4) agrees with humans 69\% (59\%) of the time in English and 62\% (60\%) in Spanish, again similar to the 63\% human-human agreement in \citet{bai2022training} and 66\% in \citet{dubois2024alpacafarm}. With further evidence in \S\ref{sec:lm-acc-on-rm-data}, we believe our LM judges reasonably reflect output quality.

We also compare our cross-lingual transfer setup to an alternative strategy, sometimes dubbed ``translate-train''~\citepia{conneau-etal-2018-xnli}, that first trains a silver target-language RM by automatically translating the source-language data and then using the silver RM for target-language alignment.
Averaged across all 30 ($=6^2-6$) cross-lingual language pairs, under \bok and judged by PaLM-2-L, our RM transfer strategy outperforms translate-train\footnote{Which we implement using Google Translate.}
(average win rate 58.8 vs. 57.5;
see \autoref{tab:summarization-bestofk} and \ref{tab:summarization-bestofk-translatedrm} for raw numbers).
RM transfer also has an efficiency advantage: to align in multiple target languages, it suffices to train one source-language RM, rather than different ones for each target language.
In \S\ref{sec:bilingual-rm}, we also explore alignment using bilingual RMs with two source languages~\citep{mulcaire-etal-2019-polyglot}, though without noticeable improvements.

\subsection{Cross-Lingual Alignment Sometimes Outperforms Monolingual Alignment} \label{sec:cross-lingual-outperforms}

Remarkably, cross-lingual reward optimization often yields an even better model than using the target-language RM.
This is validated by (1) the consistent trend when evaluated by PaLM-2-L, GPT-4, and humans, (2) their instance-level and ranking-level agreement (\S\ref{sec:transferability-as-alignment-signal}), and (3) the small confidence intervals.
This may be due to a regularization effect: the target-language RM may possess language-specific spurious artifacts, to which the target-language policy model can overfit~\citep{gao2022scaling} more than artifacts in a different language in the source-language RM.
Suppose, for example, that the target-language RM assigns higher rewards when the generation contains certain target-language words (due to bias in the RM training data). A different-language policy model is unlikely to exploit this, as it rarely generates these words, but a same-language policy model may.

\begin{figure}[t!]
    \centering
    \includegraphics[width=0.48\textwidth]{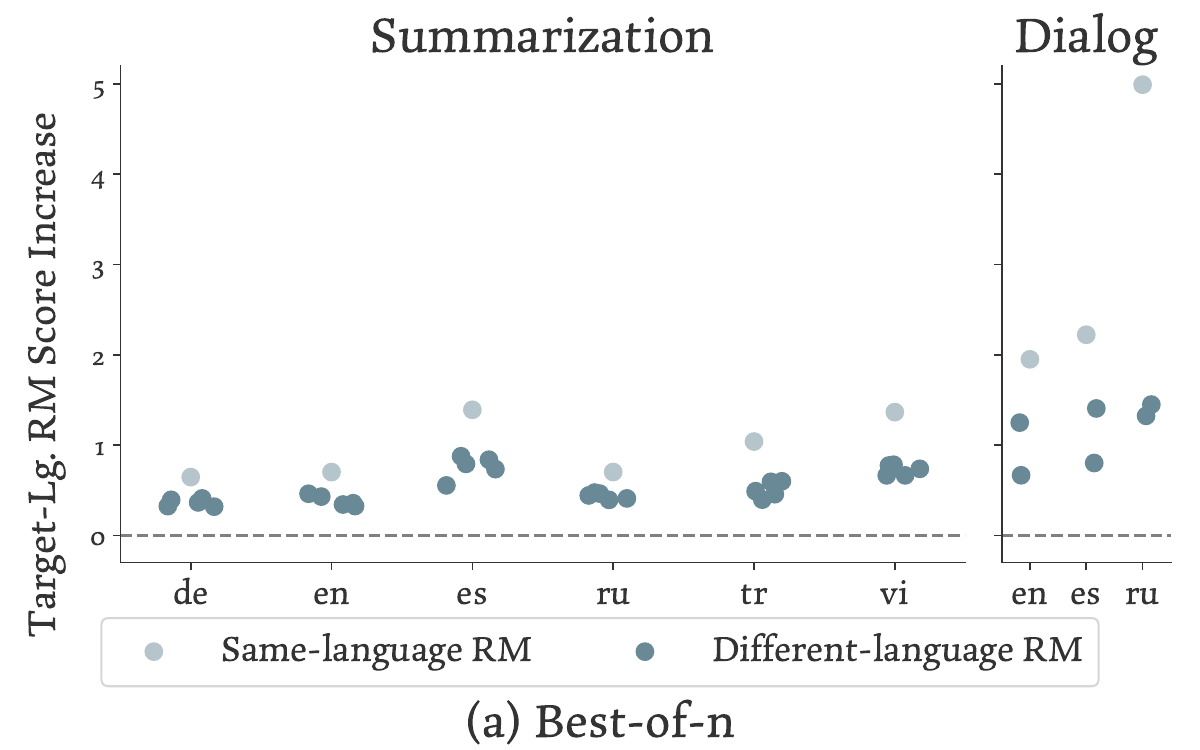} \\
    \vspace{0.3cm}
    \includegraphics[width=0.48\textwidth]{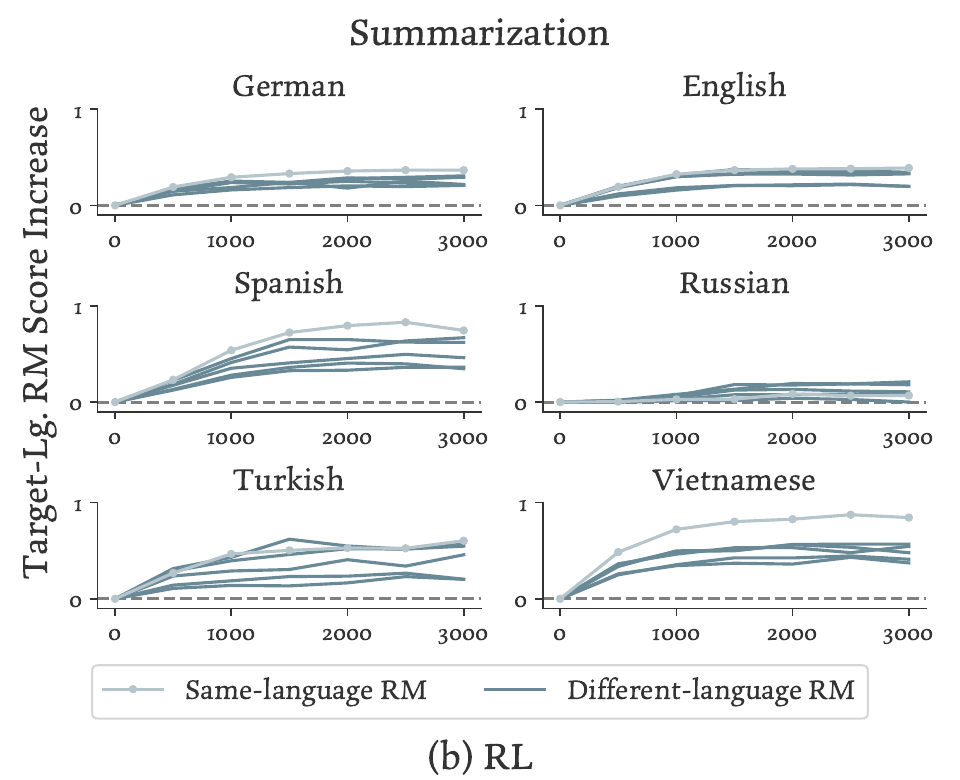}
    \caption{Cross-lingual alignment effectiveness judged by a finetuned target-language RM evaluator, measured in its score increase between the aligned model and the target-language SFT model. Each group in (a) and subplot in (b) represents one target language, and different dots/lines within each represent different source languages. RL is difficult to train for OpenAssistant~(\S\ref{sec:setup}), so we omit it here. \textbf{In most cases, the RM evaluator score improves for cross-lingually aligned models.}
    }
    \vspace{-3mm}
    \label{fig:training-dynamic-crosslingual}
\end{figure}

This hypothesis is consistent with our observed patterns.
First, there are many fewer cases of cross-lingual reward optimization outperforming the monolingual setting when measured by the finetuned target-language RM evaluator than the prompted LM evaluators~(\autoref{fig:training-dynamic-crosslingual}): under this hypothesis, the finetuned evaluator RMs would be more susceptible to such artifacts and (incorrectly) assign higher scores in the monolingual settings.
The underperformance of the translate-train baseline (\S\ref{sec:transferability-as-alignment-signal}) also provides weak evidence: in principle, a source-language RM and a source-translated-into-target-language RM should capture the same reward signal, as they are derived from the same data source, and would lead to similar downstream performance. However, the former is less susceptible to reward over-optimization due to the language mismatch, leading to better performance, though this is confounded by translation quality.

Corroborating this hypothesis, we also find that when used monolingually, the RMs behave more like a bag-of-word (BoW) model. We take each of the 6 summarization RMs and infer on the validation set of each dataset in each language~(\autoref{tab:summarization-statistics-sft}). In every setting, we fit a BoW linear regressor to predict the RM-assigned score for each instance and compute the $R^2$ across instances as a proxy for the RM's similarity to a BoW model in that setting. For each dataset, and for every source language that differs from the dataset's language, we check whether inferring using the source-language RM or the dataset-language RM results in a larger $R^2$.
The latter monolingual usage has
a higher $R^2$
(0.65 vs. 0.63), so it is more likely that the RMs overfit to lexical patterns when used in-language.

\begin{figure*}[t!]
    \centering
    \includegraphics[width=0.85\textwidth]{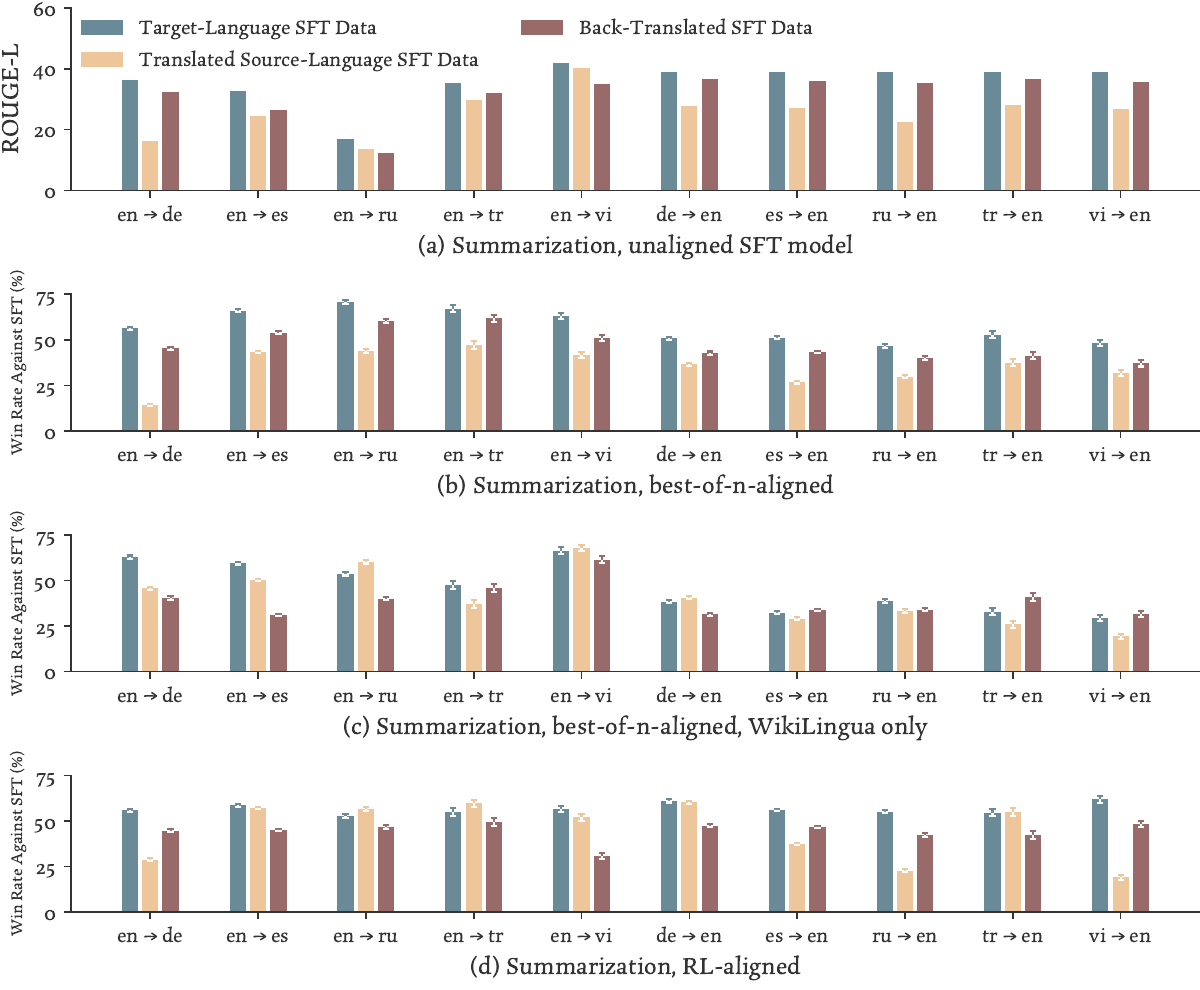}
    \caption{Cross-lingual alignment results without target-language SFT data using various strategies and on different data. 
    \textbf{Training the SFT model using data translated from another language can be helpful when aligning using RL} ((d))\textbf{, but domain match is important for \bok} ((c) and the back-translation results).}
    \vspace{-3mm}
    \label{fig:translated-sft-all}
\end{figure*}

\subsection{Cross-Lingual Alignment Without Target-Language SFT Data} \label{sec:translated-sft}

So far we assumed access to target-language SFT data since, as \S\ref{sec:method} argues, SFT data could be more easily obtained than RM data.
We now relax this assumption and instead translate the source-language SFT data into the target language using Google Translate. We investigate if it, combined with RM transfer, still enables cross-lingual alignment. As a case study, we only consider summarization and when English is the source or target language.

Using translated SFT data substantially degrades the quality of the SFT model (\autoref{fig:translated-sft-all}(a)) and the \bok-aligned LM (\autoref{fig:translated-sft-all}(b)). There are however two factors: (1) quality loss due to translation, and (2) domain/style mismatch. For (2), we note that different languages have SFT data composed of different datasets, following Seahorse~(\autoref{tab:summarization-statistics-sft}).\footnote{SFT data quantity may also be a confounder, but we consider directions both from and to English, and the degradation is substantial in both. So quantity is not the biggest factor.} And these datasets differ stylistically: for example, while XSum includes news articles, WikiLingua consists of how-to articles and with more formulaic summaries.
There would thus be a domain difference between using organic target-language SFT data vs. data translated from a different language.

To account for this, we employ round-trip back-translation, first translating the target-language SFT data into the source language and then back to the target language. This setup is not practically useful but it upper-bounds the effect of translation errors alone.
\autoref{fig:translated-sft-all}(a) shows that this bridges most of the gap, sometimes leading to models that win over the SFT model >50\% of the time.
Alternatively, we control for domain by repeating our experiments solely using WikiLingua for both SFT and RM as it is present for all languages. From \autoref{fig:translated-sft-all}(c), the gap indeed reduces, with the translated SFT models sometimes even outperforming the original, and back-translation
\iflongversion
is no longer consistently beneficial.
\else
no longer consistently helps.
\fi

Other than genre control, we also hypothesize that the gap would be smaller for RL than \bok because the RM, whose transferability we verified (\S\ref{sec:results}), intuitively plays a bigger role in the RL pipeline. Best-of-$n$, on the other hand, is more reliant on the SFT model quality, as reflected by the high resemblance between the transfer performance patterns in \autoref{fig:translated-sft-all}(b) and the SFT model quality in \autoref{fig:translated-sft-all}(a). \autoref{fig:translated-sft-all}(d) indeed shows that the translated models have little performance drop, except for cases where the former degenerates.\footnote{Which we believe is due to a lack of careful case-by-case hyperparameter tuning, which we did not perform as it would be very expensive to tune for each transfer pair.} Again, apart from the degenerate cases, back-translation is not helpful.

To summarize,\iflongversion\footnote{No pun intended.}\fi\ cross-lingual alignment could still be helpful even without target-language SFT data, though care needs to be taken when training the surrogate SFT model. While we only experimented on summarization, we believe there will be larger text diversity for dialog generation in the wild, for which this issue warrants greater attention.

\subsection{Practical Recommendations}

Our findings suggest that, for SFT, it is always beneficial to use organic target-language data, but when inaccessible, automatic translation may be a remedy, though one should be mindful of the data distribution match between the data source and the application, or relying more on RL.

\begin{figure}[t!]
    \centering
    \includegraphics[width=0.48\textwidth]{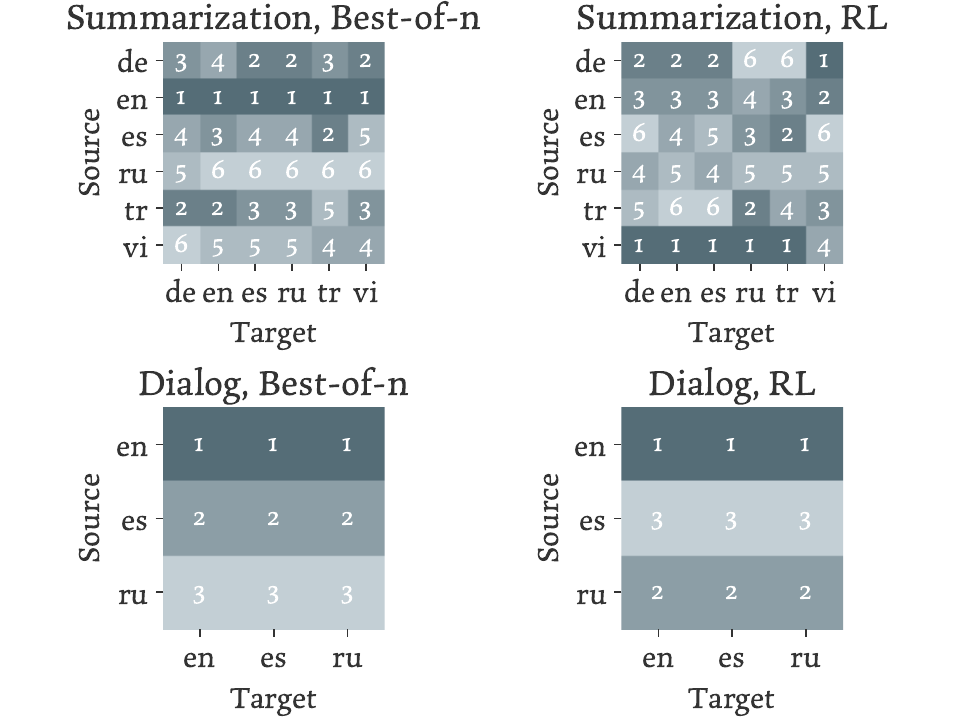}
    \caption{PaLM-2-L-judged rankings of source language effectiveness when driving alignment in different target languages.
    English is generally a good source.
    }
    \vspace{-3mm}
    \label{fig:rankings}
\end{figure}

For RM, cross-lingual transfer is often successful, but how does one select the source RM language to align in a new target language?
In \autoref{fig:rankings}, we show the source languages ranked by transfer effectiveness for each target language.
The rankings across target languages are generally stable, especially for \bok: if a source language is effective for one target language, it is usually effective for others too.
Therefore, one may select the source language by extrapolating from its performance on other target languages.
In particular, English RMs are usually the most accessible in practice. Our results show that it is a decent strategy to use them as the source: English is often a highly-ranked source language, most frequently the best, perhaps due to the relatively higher annotator quantity and quality~\citep{yu-etal-2022-beyond} or implicit modeling assumptions~\citep{dyer2019critical}.
\iflongversion
Beyond this empirical observation, we try to causally predict the pairwise transferability from various features in \S\ref{sec:analysis}, but without success.
\fi

%% file: sections/06_analysis.tex
\section{Analysis} \label{sec:analysis}

The effectiveness of cross-lingual alignment motivates us to better understand how it relates to various factors.
We show that while RM generalizability within the original reward modeling task is a prerequisite, it does not uniquely explain the downstream success.
Similarly,
we also show that the pairwise win rates (judged by PaLM-2-L unless otherwise mentioned) cannot be fully explained by, and thereby not predictable from, language features
\iflongversion
or the KL-divergence from the SFT model.
\else
or the KL-divergence from the SFT model (in \S\ref{sec:kl}).
\fi

\begin{figure*}[t!]
    \centering
    \includegraphics[width=0.9\textwidth]{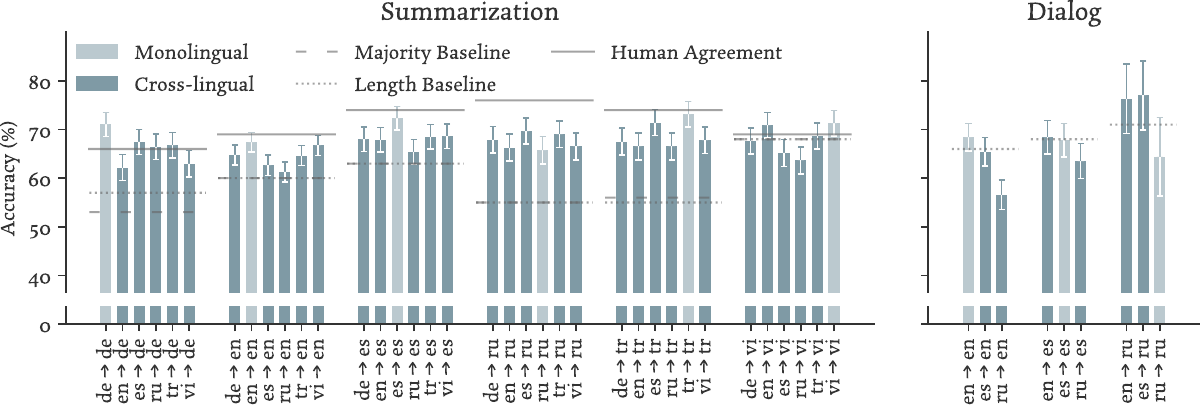}
    \caption{
    Source-language RM generalizability within the original reward modeling task and the 95\% confidence interval across validation instances. ``source$\to$target`` denotes training a source-language RM and measuring its accuracy on the target language validation data. The baselines are explained in \S\ref{sec:transferability-task-data}. Dialog generation, a pairwise task, does not have a majority baseline; the dataset authors also did not report human agreement. \textbf{RMs generally exhibit cross-lingual generalizability, exceeding the majority baseline and often the length baseline}.
    }
    \vspace{-3mm}
    \label{fig:task-data}
\end{figure*}

\subsection{Impact of RM Generalizability Within Reward Modeling}
\label{sec:transferability-task-data}

The RMs' cross-lingual utility in downstream alignment is predicated on their generalizability within the original reward modeling task, but the latter is not sufficient for the former.
So how much does this generalizability explain the alignment success?
We analyze this generalizability following the cross-lingual transfer tradition, zero-shot applying a source-language RM to the target-language validation data and computing accuracy~\citepia{wu-dredze-2019-beto,wu-dredze-2020-languages,pires-etal-2019-multilingual}.
We also consider a majority baseline and a length baseline to check if the RMs are only superficially capturing generation length~\citep{wang2023how,singhal2023long}. To compute this length baseline: for dialog generation, a pairwise task, all longer, or shorter, responses in each pair are chosen, depending on which (long or short) yields higher training set accuracy. For summarization, a pointwise task, all responses longer (or shorter) than a threshold are chosen. The direction (long or short) and the threshold are also selected using the training set.

\autoref{fig:task-data} confirms cross-lingual RM generalizability:
cross-lingual RMs often perform above the majority baseline for summarization and random performance (50\%) for dialog.
\S\ref{sec:verifying-rm-transfer-for-reward-modeling} verifies this cross-lingual generalizability with another setup.

Nevertheless, the improvements over the majority/random baselines are modest.
The dialog models even sometimes underperform the length baseline (though this does not mean the RMs only rely on length\footnote{The RMs agree with the length baseline on 72.6\% of the validation instances, higher than the baseline agreement level of 56.6\% (how often two random models at their accuracy levels agree on average), but far from full agreement.}).
Part of this is due to the high subjectivity of the reward modeling task: the RM accuracies here are near the human agreement level for Seahorse~\citep{clark-etal-2023-seahorse}, plotted in \autoref{fig:task-data}, and generally match the human agreement numbers in dialog generation work~\citep{bai2022training,dubois2024alpacafarm}.
But it is still interesting that seemingly weak RMs, like the Vietnamese RM which performs similarly to the majority baseline when used monolingually or the dialog RMs which are often surpassed by the length baseline, can achieve high cross-lingual alignment effectiveness~(\autoref{fig:palm}).

Furthermore, the results here do not match their downstream utility, regardless of whether we consider 
the quality of the RMs as measured by their in-language validation accuracy (Turkish, for example, is the best in \autoref{fig:task-data}, but not so in \autoref{fig:rankings}),
the generalizability of the RMs which we operationalize as the difference between in-language training and validation loss (or accuracy---they yield the same ranking: Russian, German, English, Turkish, Vietnamese, and Spanish, from the least amount of overfitting to the most, again different from \autoref{fig:rankings}),
or the specific pairwise transfer effectiveness (for each target language, we compare the effectiveness of source languages ranked by the reward modeling task generalizability here vs. by downstream alignment win rate;
on summarization, averaged across target languages, Kendall's $\tau=0.1$ (same with \bok or RL), indicating low ranking agreement).
Overall, while cross-lingual alignment depends on RM generalizability on the original task, other factors are at play too. 

\subsection{Impact of Language Features} \label{sec:impact-of-language-features}

Can the cross-lingual alignment performance be predicted from simple language features, such as their frequency in the pretraining corpus or typological similarity?
The summarization languages ranked by frequency in the mT5 corpus, the base model for this task, are: English, Russian, Spanish, German, Turkish, Vietnamese~\citep{xue-etal-2021-mt5}. This does not match the transfer utility ranking in \autoref{fig:rankings}. Similarly, neither does the ranking match the SFT data quantity or RM data quantity~(in \S\ref{sec:dataset-stats}).

Linguistic typology and orthography are also common predictors of cross-lingual transferability~\citepia{gerz-etal-2018-relation,K2020Cross-Lingual,muller-etal-2021-unseen}. This, however, is not the case for us either: for summarization RL, for example, English benefits from Vietnamese the most, but they belong to disparate language families. Orthography may be playing a role: Russian overall does not transfer well to other languages, and it is the only language that does not use the Latin script, but this trend is not clear. Systematically, we compute the correlation between alignment utility and WALS features of linguistic typology~\citep{wals}. For each WALS feature present for all 6 summarization languages, we divide all win rates into two groups: those between language pairs that have the same, or different, feature values. Under a
\iflongversion
one-sided unpaired
\fi
$t$-test, no feature shows statistical significance at $\alpha=0.05$ with Bonferroni correction~\citep{dunn1961multiple}.\iflongversion\footnote{Even without correction, only 4 show statistical significance at $\alpha=0.05$ out of 123: 1A, 3A, 37A, and 54A. The first two are phonological features, and the other two minor syntactic features, thus likely being spurious correlations.}
\fi\ Therefore, alignment utility does not strongly correlate with such language features.

\iflongversion

\subsection{Impact of Policy Divergence} \label{sec:kl}

\begin{figure}[t!]
    \centering
    \includegraphics[width=0.48\textwidth]{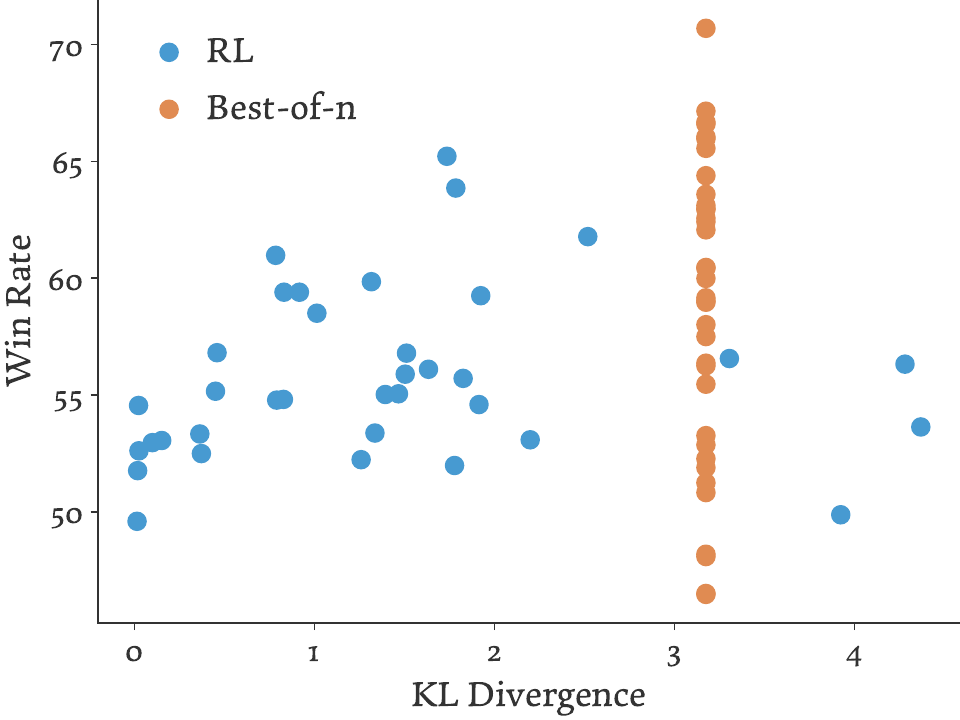}
    \caption{Win rate (PaLM-2-L-judged) vs. KL-divergence for summarization across different (source, target) language pairs.
    For \bok, we use the upper bound formula in \citet{stiennon2020learning}, \citet{beirami2024theoretical}, \emph{i.a.}, which is a function of $n$ and thus appears as a vertical line.
    \textbf{KL-divergence does not fully explain the final alignment performance}.}
    \vspace{-3mm}
    \label{fig:winrate-vs-kl}
\end{figure}

From a learning angle, it has been shown that the reward that a learned policy can obtain strongly correlates with its KL-divergence from the base (SFT) policy~\citep{bai2022training}. This could be concerning, if the model deviates from the base policy to ``hack'' the reward~\citep{gao2022scaling,coste2023reward,eisenstein2023helping}, but not if the evaluation metric is robust. As we perform human evaluation and also verified that our LM judges correlate with human judgments, this is less of a problem for us. Nevertheless, in \autoref{fig:winrate-vs-kl}, we plot the correlation between the win rates and the KL-divergence of the aligned models. There is not a clear correlation, and hence we do not observe reward over-optimization.

\fi

%% file: sections/08_related_work.tex
\section{Related Work}

\paragraph{Zero-shot cross-lingual transfer.} There is a long line of research on cross-lingual representation generalizability, such as with sentence embeddings~\citep{conneau-etal-2018-xnli} or more recently, LMs~\citep{wu-dredze-2019-beto,wu-dredze-2020-languages,pires-etal-2019-multilingual,siddhant2020evaluating}. Commonly, a multilingual LM~\citepia{devlin-etal-2019-bert,conneau2019cross,conneau-etal-2020-unsupervised} is finetuned on a task in a source language and evaluated on the task's test set in a different language. This is generally effective.
Our RM transfer setup can be viewed under this framework, but we go further and show that this generalizability is useful for \emph{downstream} tasks, in our case alignment.
\citet{shaham2024multilingual} and \citet{chirkova2024zeroshot} are close to us in studying cross-lingual generalizability in alignment, but only focusing on SFT and only using translated data.

\paragraph{Multilingual Alignment.}
For SFT, it is common to assemble existing multilingual task datasets into instruction datasets~\citep{muennighoff-etal-2023-crosslingual,asai2023buffet,ahuja-etal-2023-mega}.
Some have directly collected SFT data for non-English languages, either on a per-language basis~\citepia{zhang2023chinese,xu2023cvalues,instructionwild} or multilingually~\citep{zhao2024inthewildchat,singh2024aya}, though this can be expensive.
Past work has also used automatic translation for SFT~\citepia{li2023bactrianx,lai-etal-2023-okapi,shaham2024multilingual} and RM data~\citep{lai-etal-2023-okapi,shen2024language}.
We also use translation for SFT, but showed that cross-lingual transfer outperforms translation for RM.

%% file: sections/09_conclusion.tex
\section{Conclusion}

We showed through two different tasks that we can perform alignment using a different-language RM. Surprisingly, we find this to be sometimes more effective than using a same-language RM. We also identified issues and remedies when we dispense with target-language SFT data.
We hope our findings can motivate future work to build better LMs for more languages.
Adapting our RM transfer setup to other settings such as domain generalization would also be exciting future directions.

%% file: sections/99_appendix.tex
\section{Dataset Details and Statistics} \label{sec:dataset-stats}

We report dataset statistics in \autoref{tab:summarization-statistics-sft}, \ref{tab:summarization-statistics-rm}, \ref{tab:openassistant-statistics-sft}, and \ref{tab:openassistant-statistics-rm}.
We reuse the SFT data for reward optimization (for both training and evaluation for RL, and for only evaluation for \bok since it does not have a training stage), but only the input $x$, without reference generations $y$.

The summarization SFT datasets, reported in \autoref{tab:summarization-statistics-sft},
are the original data sources of Seahorse, which we take from the GEM release~\citep{gehrmann-etal-2021-gem}.
They are evenly mixed at the instance level for both SFT training and RL training.
For evaluation of the aligned model, we macro-average the per-dataset metrics (e.g., win rate) for a language-level score.
Because the Seahorse dataset was created using the validation and test instances of the original summarization datasets, to be clean, we exclude the Seahorse training instances from these splits when performing SFT and reward optimization. OpenAssistant does not have this issue and has clean split separations.
The Seahorse summaries are human-rated along six axes, and we only use the sixth axis for our pointwise reward as it encapsulates previous axes~\citep{clark-etal-2023-seahorse}.
We limit the maximum length of model inputs to 1,024 tokens and outputs to 512 tokens.
See also \S\ref{sec:task-instructions} for instructions we attach to the dataset instances during training and inference.

\begin{table}[t!]
    \centering
    \begin{tabular}{@{\hspace{4pt}}c@{\hspace{4pt}}c|cc@{\hspace{4pt}}}
        \toprule
        & & Train & Validation \\
        \midrule
        \multirow{2}{*}{German} & MLSum & 220748 & \phantom{0}8932 \\
        & WikiLingua & \phantom{0}40839 & \phantom{0}3699 \\
        \midrule
        \multirow{3}{*}{English} & XSum & \phantom{0}23206 & \phantom{00}642 \\
        & XL-Sum & 306522 & \phantom{0}9690 \\
        & WikiLingua & \phantom{0}99020 & 12021 \\
        \midrule
        \multirow{3}{*}{Spanish} & XL-Sum & \phantom{0}38110 & \phantom{0}3170 \\
        & MLSum & 259888 & \phantom{0}8374 \\
        & WikiLingua & \phantom{0}79212 & \phantom{0}9730 \\
        \midrule
        \multirow{2}{*}{Russian} & XL-Sum & \phantom{0}62243 & \phantom{0}5492 \\
        & WikiLingua & \phantom{0}37028 & \phantom{0}3209 \\
        \midrule
        \multirow{2}{*}{Turkish} & XL-Sum & \phantom{0}27176 & \phantom{0}1953 \\
        & WikiLingua & \phantom{00}3148 & \phantom{00}194 \\
        \midrule
        \multirow{2}{*}{Vietnamese} & XL-Sum & \phantom{0}32111 & \phantom{0}2341 \\
        & WikiLingua & \phantom{0}13707 & \phantom{00}679 \\
        \bottomrule        
    \end{tabular}
    \caption{\label{tab:summarization-statistics-sft}
    Number of summarization instances for the SFT and reward optimization stages. The datasets are taken from the GEM release~\citep{gehrmann-etal-2021-gem} and with certain validation instances removed~(\S\ref{sec:dataset-stats}).
    }
\end{table}

\begin{table}[t!]
    \centering
    \begin{tabular}{c|cc}
        \toprule
        & Train & Validation \\
        \midrule
        German & \phantom{0}8389 & 1250 \\
        English & 14031 & 2071 \\
        Spanish & \phantom{0}8741 & 1310 \\
        Russian & \phantom{0}7679 & 1112 \\
        Turkish & \phantom{0}7855 & 1096 \\
        Vietnamese & \phantom{0}7844 & 1166 \\
        \bottomrule
    \end{tabular}
    \caption{\label{tab:summarization-statistics-rm}
    Number of summarization instances for reward modeling.
    }
\end{table}

\begin{table}[t!]
    \centering
    \begin{tabular}{c|cc}
        \toprule
        & Train & Validation \\
        \midrule
        English & 8898 & 472 \\
        Spanish & 5681 & 311 \\
        Russian & 1884 & \phantom{0}99 \\
        \bottomrule
    \end{tabular}
    \caption{\label{tab:openassistant-statistics-sft}
    Number of dialog generation instances for the SFT and reward optimization stages.
    }
\end{table}

\begin{table}[t!]
    \centering
    \begin{tabular}{c|cc}
        \toprule
        & Train & Validation \\
        \midrule
        English & 22076 & 1026 \\
        Spanish & 13714 & \phantom{0}699 \\
        Russian & \phantom{0}2627 & \phantom{0}135 \\
        \bottomrule
    \end{tabular}
    \caption{\label{tab:openassistant-statistics-rm}
    Number of dialog generation instances for reward modeling.
    }
\end{table}

\section{Training Details} \label{sec:training-details}

\paragraph{SFT.} The model is trained using Adafactor~\citep{adafactor} with a constant learning rate at $10^{-3}$ for summarization and $10^{-5}$ for dialog generation, batch size 32, and dropout 0.1. We perform checkpoint selection using validation ROUGE-L score~\citep{lin-2004-rouge}.

\paragraph{RM.} The model is trained using Adafactor with a constant learning rate at $10^{-4}$ after 1,000 linear warm-up steps, batch size 32, and dropout 0.1. We perform checkpoint selection using validation loss.

\paragraph{RL.} We use PPO for RL training with a constant learning rate at $10^{-4}$, batch size 32, for 3,000 steps for summarization and 2,500 steps for dialog generation. The value model has 1,000 linear warm-up steps and we only start training the policy model after 2,000 steps elapse. We set the regularization coefficient at $\beta=0.1$.

\paragraph{Best-of-$n$.} We use $n=64$.

\begin{table*}[t!]
    \centering
    \begin{tabular}{c|c|cccccc}
        \toprule
        && De & En & Es & Ru & Tr & Vi \\
        \midrule
        \multirow{2}{*}{Summarization} & Acc. & 73.5\% & 73.0\% & 73.2\% & 73.7\% & 73.6\% & 78.2\% \\
        & $N$ & 306 & 1672 & 295 & 255 & 720 & 349 \\
        \midrule
        \multirow{2}{*}{Dialog} & Acc. & -- & 72.0\% & 70.8\% & 73.3\% & -- & -- \\
        & $N$ & -- & 472 & 311 & 99 & -- & -- \\
        \bottomrule
    \end{tabular}
    \caption{\label{tab:lm-acc-on-rm-data}
    The accuracy of evaluating the PaLM-2-L judge on the RM validation data. We also report the number of comparisons based on which the accuracy is calculated.
    }
\end{table*}

\begin{figure*}[t!]
    \centering
    \includegraphics[width=\textwidth]{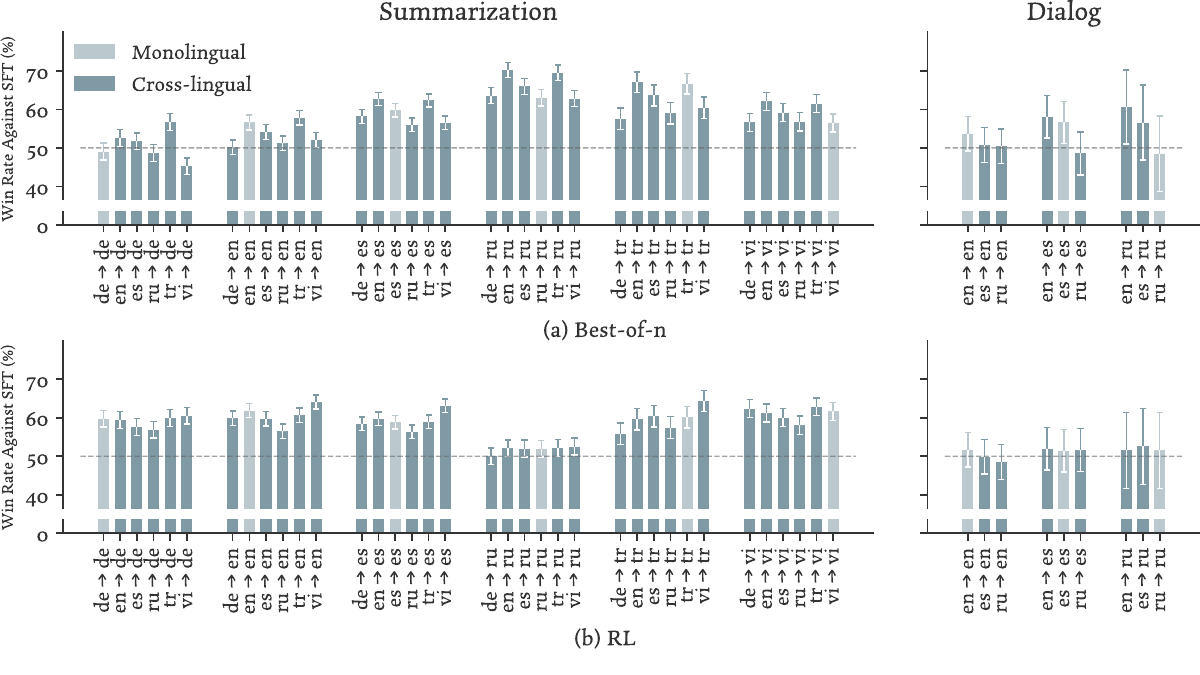}
    \vspace{-1cm}
    \caption{Alignment effectiveness, compared to the target-language SFT model judged by GPT-4, and the 95\% confidence interval across validation instances. ``source$\to$target`` denotes a source-language RM driving alignment in the target language. \textbf{Cross-lingual alignment is generally effective, sometimes outperforming monolingual alignment}. RL is hard to train for OpenAssistant, in line with what its authors found~\citep{kopf2023openassistant}.}
    \label{fig:gpt4}
\end{figure*}

\section{LM Judge Accuracy on Ground-truth Reward Modeling Data} \label{sec:lm-acc-on-rm-data}

We verify the validity of using LM as a judge for our tasks by computing its accuracy on the validation splits of the RM datasets we used. We only consider PaLM-2-L as a case study. For OpenAssistant, a pairwise dataset, we simply check if the RM ranks the candidate generations correctly according to human preference. For Seahorse, a pointwise dataset, we group summaries for the same source document, and for each summary pair in such groups, we compute the ranking correctness.

We show the results in Table~\ref{tab:lm-acc-on-rm-data}. The accuracies generally match the human agreement in Seahorse~\citep{clark-etal-2023-seahorse}, and while human agreement was not reported in OpenAssistant, they generally match the human agreement numbers in past work on dialog generation~\citep{bai2022training,dubois2024alpacafarm} too (see \S\ref{sec:transferability-as-alignment-signal} for reference human agreement numbers). Taken together with the LM judges' agreement with human evaluation~(\S\ref{sec:transferability-as-alignment-signal}), we believe it is valid to use a LM to assess the generation quality in our setup.

\section{GPT-4 as a Judge Results} \label{sec:gpt4-results}

In this section, we present the alignment evaluation results as judged by GPT-4, specifically the \texttt{gpt-4-0125-preview} model. Due to its high cost, we cap the number of evaluation instances for each dataset at 1,000 (i.e., for each row of \autoref{tab:summarization-statistics-sft} and \ref{tab:openassistant-statistics-sft}). The results are shown in \autoref{fig:gpt4}. We observe the same trends as in \S\ref{sec:transferability-as-alignment-signal}, where cross-lingual reward optimization is generally effective, sometimes even more so than when done monolingually. Compared to PaLM-2-L, the two LMs agree on 72\% of the instances in English and 71\% in Spanish for summarization, and 75\% and 73\% for these languages for dialog. These are higher than the baseline human-human agreement numbers in \S\ref{sec:transferability-as-alignment-signal}. This shows a sign of homogeneity between LM judges, but also confirms their reliability.

\section{Verifying RM Transfer for Reward Modeling}
\label{sec:verifying-rm-transfer-for-reward-modeling}

\begin{figure}[t!]
    \centering
    \includegraphics[width=0.45\textwidth]{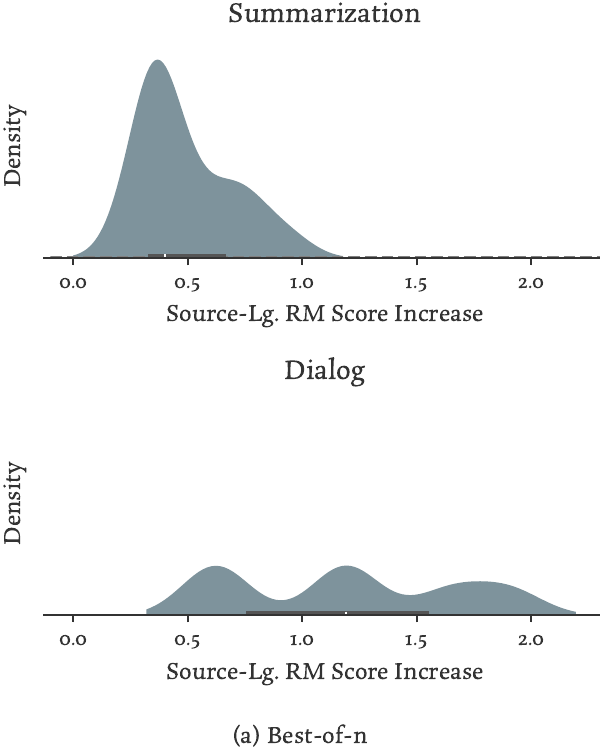} \\
    \vspace{0.5cm}
    \includegraphics[width=0.45\textwidth]{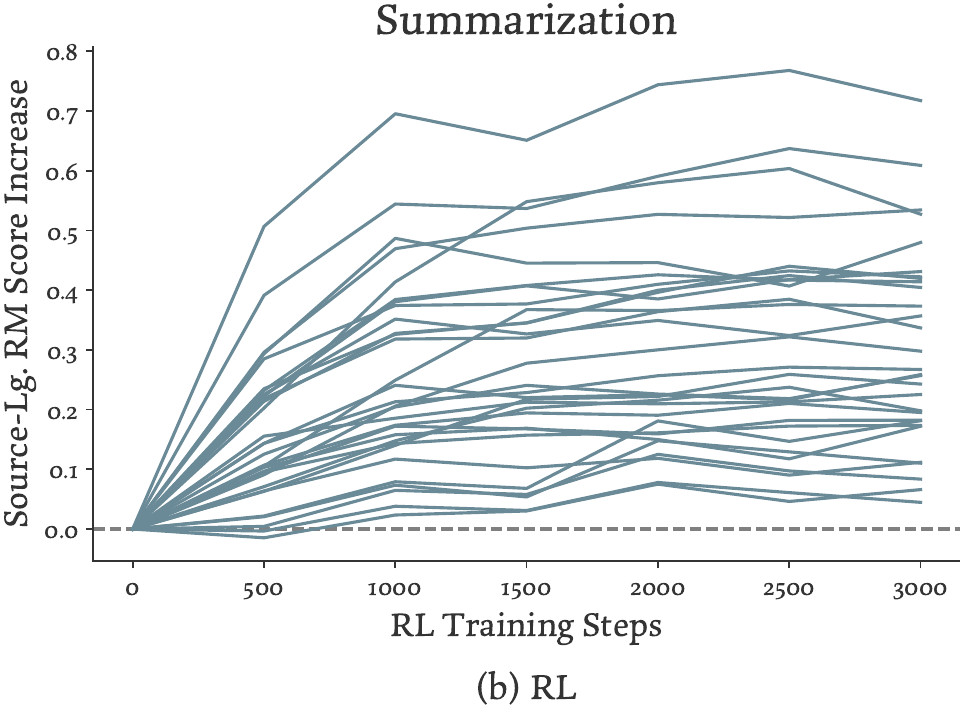}
    \caption{Source-language RM generalizability evaluated by increases in scores they assign to target-language generations after monolingual target-language alignment (\bok or RL). We show all (source, target) language pairs where the two languages differ as density in (a) and lines in (b). RL is difficult to train for OpenAssistant (\S\ref{sec:setup}), so we omit it here, since the assumption that the RL'ed model is better would not hold. \textbf{In most cases, the source-language RM assigns a higher score (\textgreater0 increase) to aligned models, demonstrating cross-lingual RM generalizability.}}
    \label{fig:training-dynamic-rmeval}
\end{figure}

In \S\ref{sec:transferability-task-data}, we observed RM generalizability on the original reward modeling task, which would be a necessary condition for successful downstream cross-lingual alignment. There, we showed that the source-language RMs assign higher scores to better target-language generations than worse ones. Here, we consider an alternative setup to study the same problem: instead of relying on existing RM datasets for the better and worse generations, we take generations from monolingually-aligned models as better ones than those from unaligned (i.e., SFT) models. The assumption here is that monolingual alignment improves model quality, which is indeed the case as illustrated in \autoref{fig:palm} and \ref{fig:gpt4}. Like in \S\ref{sec:transferability-task-data}, we indeed see from \autoref{fig:training-dynamic-rmeval} that source-language RMs assign higher scores to monolingually-aligned models than unaligned SFT models.
Under RL, this score difference also increases throughout training.
These results confirm the RMs' cross-lingual generalizability within the reward modeling task.

\section{Alignment Using Bilingual RMs} \label{sec:bilingual-rm}

\begin{figure*}[t!]
    \centering
    \includegraphics[width=\textwidth]{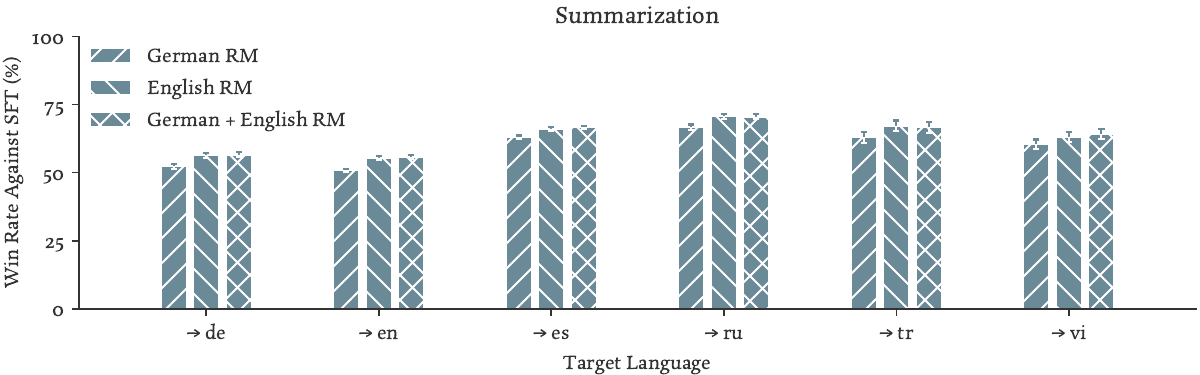}
    \caption{Alignment performance, measured in the win rate against the monolingual target-language SFT model, when alignment is driven by a German RM, an English RM, or a bilingual German + English RM. \textbf{The bilingual RM does not yield a noticeable improvement}.}
    \label{fig:bilingual-rm}
\end{figure*}

Seeing the benefit of cross-lingual RM transferability in \S\ref{sec:results}, we hypothesize that bilingual RMs could bring further improvements since the resulting reward could be encouraged to be more language agnostic~\citep{mulcaire-etal-2019-polyglot}. It would be computationally expensive to experiment with all possible language configurations (there would be a cubic number of them with pairwise sources), so, for simplicity, we take the best-performing source languages under the summarization \bok setup as judged by PaLM-2-L, English and German (\autoref{fig:rankings}), and see if a bilingual RM based on them would lead to further performance improvement. Specifically, we first train a bilingual SFT model by pooling the SFT data for both languages, and similarly for the RM, which initializes from this bilingual SFT model.

\autoref{fig:bilingual-rm} does not show an improvement from the bilingual RM, which always achieves similar performance to the English RM, the better of the two monolingual RMs. Nevertheless, if this trend holds consistently, that the bilingual RM matches the performance of the better monolingual RM, this could be useful as an alternative to having to perform source language selection.
We leave a more systematic validation of this phenomenon to future work.

\iflongversion
\else 

\section{Impact of Policy Divergence} \label{sec:kl}

\begin{figure}[t!]
    \centering
    \includegraphics[width=0.48\textwidth]{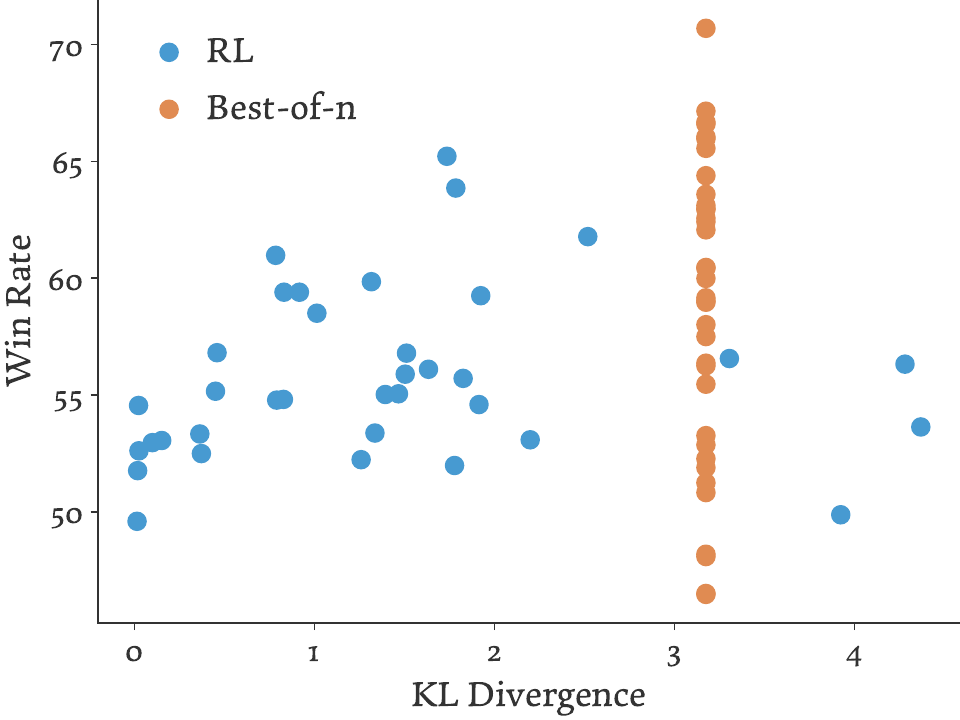}
    \caption{Win rate (PaLM-2-L-judged) vs. KL-divergence for the summarization task across different (source, target) language pairs.
    For \bok, we use the upper bound formula in \citet{stiennon2020learning}, \citet{beirami2024theoretical}, \emph{i.a.}, which is a function of $n$ and thus appears as a vertical line.
    \textbf{KL-divergence does not fully explain the final alignment performance}.}
    \vspace{-3mm}
    \label{fig:winrate-vs-kl}
\end{figure}

From a learning angle, it has been shown that the reward that a learned policy can obtain strongly correlates with its KL-divergence from the base (SFT) policy~\citep{bai2022training}. This could be concerning, if the model deviates from the base policy to ``hack'' the reward~\citep{gao2022scaling,coste2023reward,eisenstein2023helping}, but not if the evaluation metric is robust. As we perform human evaluation and also verified that our LM judges correlate with human judgments, this is less of a problem for us. Nevertheless, in \autoref{fig:winrate-vs-kl}, we plot the correlation between the win rates and the KL-divergence of the aligned models. There is not a clear correlation, and hence we do not observe reward over-optimization.

\fi

\section{Prompts}

In this section, we list all the prompts we used.

\subsection{Task Instructions} \label{sec:task-instructions}

We prepend the following task-specific instructions to inputs for SFT and reward optimization. All occurrences of \texttt{[LANGUAGE]} are substituted with the target language. The RM stage does not include such prompts, where we simply concatenate the texts with delimiters.

Summarization: \texttt{Summarize the following text in [LANGUAGE]: }

Dialog generation: \texttt{You are given a dialog between a human and an assistant in  [LANGUAGE]. Please write one turn of the assistant side in [LANGUAGE].\textbackslash n\textbackslash n''}

\subsection{Evaluation Prompts} \label{sec:lm-eval-prompts}

We use the following prompts to elicit pairwise generation judgments for both human and LM judge evaluation. All occurrences of \texttt{[LANGUAGE]}, \texttt{[INPUT]}, \texttt{[GENERATION1]}, and \texttt{[GENERATION2]} are substituted with the respective content. For both tasks, we compare the probability of the tokens ``\texttt{1}'' and ``\texttt{2}''. To control for the positional bias of LMs~\citep{wang2023large,pezeshkpour2023large,zheng2023large} and potentially of our human annotators, we randomly shuffle the two generations for human evaluation and the GPT-4 judge. For the PaLM-2 judge for which we have probability access, we prompt the LM judge twice with both orderings of the generations and compute the accuracy by averaging the probabilities of the ``\texttt{1}'' and ``\texttt{2}'' tokens.

\paragraph{Summarization.} This prompt is adapted from the one in \citet{lee2023rlaif}.

\begin{verbatim}
A good summary is a shorter piece of text
that has the essence of the original. It
tries to accomplish the same purpose and
conveys the key information from the
original post. Below we define four
evaluation axes for summary quality:
coherence, accuracy, coverage, and overall
quality.

Coherence: This axis answers the question
“how coherent is the summary on its own?”
A summary is coherent if it's easy to
understand when read on its own and free of
English errors. A summary is not coherent
if it's difficult to understand what the
summary is trying to say. Generally, it's
more important that the summary is
understandable than it being free of
grammar errors.

Accuracy: This axis answers the question
“does the factual information in the
summary accurately match the post?” A
summary is accurate if it doesn't say
things that aren't in the article, it
doesn't mix up people, and generally is
not misleading.

Coverage: This axis answers the question
“how well does the summary cover the
important information in the post?” A
summary has good coverage if it mentions
the main information from the post that's
important to understand the situation
described in the post. A summary has poor
coverage if someone reading only the
summary would be missing several important
pieces of information about the situation
in the post. A summary with good coverage
should also match the purpose of the
original post (e.g. to ask for advice).

Overall quality: This axis answers the
question “how good is the summary overall
at representing the post?” This can
encompass all of the above axes of quality,
as well as others you feel are important.
If it's hard to find ways to make the
summary better, the overall quality is
good. If there are lots of different ways
the summary can be made better, the overall
quality is bad.

You are an expert summary rater and are
knowledgeable in [LANGUAGE]. Given a
piece of text in [LANGUAGE] and two of its
possible summaries, also in [LANGUAGE],
output 1 or 2 to indicate which summary
best adheres to coherence, accuracy,
coverage, and overall quality as defined
above.

Text - [INPUT]
Summary 1 - [GENERATION1]
Summary 2 - [GENERATION2]

Preferred Summary=
\end{verbatim}

\paragraph{Dialog Generation}

This prompt is adapted from the one in \citet{alpaca_eval}.

\begin{verbatim}
You are a helpful assistant, that ranks
models by the quality of their answers.
You are also knowledgeable in [LANGUAGE].

I want you to create a leaderboard of
different large-language models. To do
so, I will give you the instructions
(prompts) given to the models, and the
responses of two models. Please rank the
models based on which response would be
preferred by humans. All inputs are
python dictionaries.

Here is the prompt, in [LANGUAGE]:
{  
    "instruction": """[INPUT]""",
}

Here are the outputs of the models, also
in [LANGUAGE]:
[   
    {  
        "model": "model_1",
        "answer": """[GENERATION1]"""
    },
    {  
        "model": "model_2",
        "answer": """[GENERATION2]"""
    }
]

Respond 1 or 2 to indicate the better
output. Please provide the ranking that
the majority of humans would give.

Better output=
\end{verbatim}

\section{Raw Results} \label{sec:raw-results}

We show the raw numerical results that correspond to our plots in \autoref{tab:summarization-bestofk} to \ref{tab:bilingual-rm}.

\begin{table}[t!]
    \centering
    \begin{tabular}{@{\hspace{4pt}}c@{\hspace{4pt}}|@{\hspace{6pt}}c@{\hspace{6pt}}c@{\hspace{6pt}}c@{\hspace{6pt}}c@{\hspace{6pt}}c@{\hspace{6pt}}c@{\hspace{4pt}}}
        \toprule
        Src \textbackslash\ Tgt & De & En & Es & Ru & Tr & Vi \\
        \midrule
        De & 52.3 & 50.8 & 63.0 & 66.7 & 63.0 & 60.4 \\
        En & 56.4 & 55.5 & 66.1 & 70.7 & 67.2 & 63.1 \\
        Es & 51.9 & 51.2 & 62.4 & 66.0 & 64.4 & 57.5 \\
        Ru & 48.1 & 46.5 & 59.2 & 63.6 & 59.0 & 56.3 \\
        Tr & 53.3 & 52.9 & 62.6 & 66.6 & 60.4 & 59.0 \\
        Vi & 46.5 & 48.2 & 60.0 & 65.6 & 62.1 & 58.0 \\
        \bottomrule
    \end{tabular}
    \caption{\label{tab:summarization-bestofk}
    Cross-lingual alignment results using \textbf{\bok} with $n=64$, for the \textbf{summarization} task, measured in win rate (\%) against the target-language SFT model as judged by \textbf{PaLM-2-L}~(\autoref{fig:palm}).
    }
\end{table}

\begin{table}[t!]
    \centering
    \begin{tabular}{@{\hspace{4pt}}c@{\hspace{4pt}}|@{\hspace{6pt}}c@{\hspace{6pt}}c@{\hspace{6pt}}c@{\hspace{4pt}}}
        \toprule
        Src \textbackslash\ Tgt & En & Es & Ru \\
        \midrule
        En & 62.9 & 65.0 & 59.6 \\
        Es & 59.1 & 62.4 & 57.6 \\
        Ru & 53.4 & 54.3 & 52.5 \\
        \bottomrule
    \end{tabular}
    \caption{\label{tab:openassistant-bestofk}
    Cross-lingual alignment results using \textbf{\bok} with $n=64$, for the \textbf{dialog generation} task, measured in win rate (\%) against the target-language SFT model as judged by \textbf{PaLM-2-L}~(\autoref{fig:palm}).
    }
\end{table}

\begin{table}[t!]
    \centering
    \begin{tabular}{@{\hspace{4pt}}c@{\hspace{4pt}}|@{\hspace{6pt}}c@{\hspace{6pt}}c@{\hspace{6pt}}c@{\hspace{6pt}}c@{\hspace{6pt}}c@{\hspace{6pt}}c@{\hspace{4pt}}}
        \toprule
        Src \textbackslash\ Tgt & De & En & Es & Ru & Tr & Vi \\
        \midrule
        De & 59.4 & 61.0 & 59.4 & 49.6 & 52.5 & 59.3 \\
        En & 55.9 & 59.9 & 58.5 & 52.6 & 54.8 & 56.6 \\
        Es & 52.0 & 56.1 & 56.8 & 53.0 & 55.0 & 49.9 \\
        Ru & 54.8 & 55.2 & 56.8 & 51.8 & 53.3 & 52.2 \\
        Tr & 53.1 & 54.6 & 55.7 & 53.1 & 53.4 & 56.3 \\
        Vi & 63.9 & 61.8 & 65.2 & 54.6 & 55.1 & 53.6 \\
        \bottomrule
    \end{tabular}
    \caption{\label{tab:summarization-rlhf}
    Cross-lingual alignment results using \textbf{RL}, for the \textbf{summarization} task, measured in win rate (\%) against the target-language SFT model as judged by \textbf{PaLM-2-L}~(\autoref{fig:palm}).
    }
\end{table}

\begin{table}[t!]
    \centering
    \begin{tabular}{@{\hspace{4pt}}c@{\hspace{4pt}}|@{\hspace{6pt}}c@{\hspace{6pt}}c@{\hspace{6pt}}c@{\hspace{4pt}}}
        \toprule
        Src \textbackslash\ Tgt & En & Es & Ru \\
        \midrule
        En & 53.1 & 54.5 & 53.5 \\
        Es & 49.9 & 51.1 & 47.5 \\
        Ru & 51.2 & 52.7 & 52.5 \\
        \bottomrule
    \end{tabular}
    \caption{\label{tab:openassistant-rlhf}
    Cross-lingual alignment results using \textbf{RL}, for the \textbf{dialog generation} task, measured in win rate (\%) against the target-language SFT model as judged by \textbf{PaLM-2-L}~(\autoref{fig:palm}).
    }
\end{table}

\begin{table}[t!]
    \centering
    \begin{tabular}{@{\hspace{4pt}}c@{\hspace{4pt}}|@{\hspace{6pt}}c@{\hspace{6pt}}c@{\hspace{6pt}}c@{\hspace{6pt}}c@{\hspace{6pt}}c@{\hspace{6pt}}c@{\hspace{4pt}}}
        \toprule
        Src \textbackslash\ Tgt & De & En & Es & Ru & Tr & Vi \\
        \midrule
        De & 49.0 & 50.2 & 58.2 & 63.6 & 57.6 & 56.6 \\
        En & 52.6 & 56.6 & 62.7 & 70.2 & 67.0 & 62.1 \\
        Es & 51.7 & 54.1 & 59.8 & 65.9 & 63.6 & 59.2 \\
        Ru & 48.7 & 51.2 & 56.0 & 63.0 & 59.0 & 56.8 \\
        Tr & 56.7 & 57.8 & 62.3 & 69.5 & 66.6 & 61.5 \\
        Vi & 45.2 & 52.1 & 56.6 & 62.8 & 60.5 & 56.5 \\
        \bottomrule
    \end{tabular}
    \caption{\label{tab:summarization-bestofk-gpt4}
    Cross-lingual alignment results using \textbf{\bok} with $n=64$, for the \textbf{summarization} task, measured in win rate (\%) against the target-language SFT model as judged by \textbf{GPT-4}~(\autoref{fig:gpt4}).
    }
\end{table}

\begin{table}[t!]
    \centering
    \begin{tabular}{@{\hspace{4pt}}c@{\hspace{4pt}}|@{\hspace{6pt}}c@{\hspace{6pt}}c@{\hspace{6pt}}c@{\hspace{4pt}}}
        \toprule
        Src \textbackslash\ Tgt & En & Es & Ru \\
        \midrule
        En & 53.7 & 58.0 & 60.6 \\
        Es & 50.7 & 56.6 & 56.6 \\
        Ru & 50.4 & 48.6 & 48.5 \\
        \bottomrule
    \end{tabular}
    \caption{\label{tab:openassistant-bestofk-gpt4}
    Cross-lingual alignment results using \textbf{\bok} with $n=64$, for the \textbf{dialog generation} task, measured in win rate (\%) against the target-language SFT model as judged by \textbf{GPT-4}~(\autoref{fig:gpt4}).
    }
\end{table}

\begin{table}[t!]
    \centering
    \begin{tabular}{@{\hspace{4pt}}c@{\hspace{4pt}}|@{\hspace{6pt}}c@{\hspace{6pt}}c@{\hspace{6pt}}c@{\hspace{6pt}}c@{\hspace{6pt}}c@{\hspace{6pt}}c@{\hspace{4pt}}}
        \toprule
        Src \textbackslash\ Tgt & De & En & Es & Ru & Tr & Vi \\
        \midrule
        De & 59.8 & 59.9 & 58.4 & 50.0 & 55.8 & 62.4 \\
        En & 59.4 & 61.8 & 59.7 & 52.1 & 59.6 & 61.2 \\
        Es & 57.6 & 59.7 & 58.8 & 52.0 & 60.4 & 60.1 \\
        Ru & 56.9 & 56.5 & 56.4 & 52.0 & 57.4 & 58.0 \\
        Tr & 59.9 & 60.7 & 59.0 & 52.2 & 60.1 & 62.8 \\
        Vi & 60.5 & 64.1 & 63.1 & 52.5 & 64.4 & 61.6 \\
        \bottomrule
    \end{tabular}
    \caption{\label{tab:summarization-rlhf-gpt4}
    Cross-lingual alignment results using \textbf{RL}, for the \textbf{summarization} task, measured in win rate (\%) against the target-language SFT model as judged by \textbf{GPT-4}~(\autoref{fig:gpt4}).
    }
\end{table}

\begin{table}[t!]
    \centering
    \begin{tabular}{@{\hspace{4pt}}c@{\hspace{4pt}}|@{\hspace{6pt}}c@{\hspace{6pt}}c@{\hspace{6pt}}c@{\hspace{4pt}}}
        \toprule
        Src \textbackslash\ Tgt & En & Es & Ru \\
        \midrule
        En & 51.7 & 51.9 & 51.5 \\
        Es & 49.9 & 51.5 & 52.5 \\
        Ru & 48.5 & 51.6 & 51.5 \\
        \bottomrule
    \end{tabular}
    \caption{\label{tab:openassistant-rlhf-gpt4}
    Cross-lingual alignment results using \textbf{RL}, for the \textbf{dialog generation} task, measured in win rate (\%) against the target-language SFT model as judged by \textbf{GPT-4}~(\autoref{fig:gpt4}).
    }
\end{table}

\begin{table}[t!]
    \centering
    \begin{tabular}{@{\hspace{4pt}}c@{\hspace{4pt}}|@{\hspace{6pt}}c@{\hspace{6pt}}c@{\hspace{4pt}}}
        \toprule
        Src \textbackslash\ Tgt & En & Es \\
        \midrule
        De & 61.0 & 64.0 \\
        En & 60.9 & 67.4 \\
        Es & 62.6 & 69.0 \\
        Ru & 51.9 & 63.4 \\
        Tr & 61.8 & 66.3 \\
        Vi & 52.3 & 61.2 \\
        \bottomrule
    \end{tabular}
    \caption{\label{tab:summarization-bok-human}
    Cross-lingual alignment results using \textbf{\bok}, for the \textbf{summarization} task, measured in win rate (\%) against the target-language SFT model as judged by \textbf{human evaluators}~(\autoref{fig:humans}).
    }
\end{table}

\begin{table}[t!]
    \centering
    \begin{tabular}{@{\hspace{4pt}}c@{\hspace{4pt}}|@{\hspace{6pt}}c@{\hspace{6pt}}c@{\hspace{4pt}}}
        \toprule
        Src \textbackslash\ Tgt & En & Es \\
        \midrule
        De & 64.4 & 64.2 \\
        En & 61.4 & 65.9 \\
        Es & 58.7 & 62.7 \\
        Ru & 61.9 & 60.6 \\
        Tr & 63.3 & 64.9 \\
        Vi & 66.2 & 64.7 \\
        \bottomrule
    \end{tabular}
    \caption{\label{tab:summarization-rlhf-human}
    Cross-lingual alignment results using \textbf{RL}, for the \textbf{summarization} task, measured in win rate (\%) against the target-language SFT model as judged by \textbf{human evaluators}~(\autoref{fig:humans}).
    }
\end{table}

\begin{table}[t!]
    \centering
    \begin{tabular}{@{\hspace{4pt}}c@{\hspace{4pt}}|@{\hspace{6pt}}c@{\hspace{6pt}}c@{\hspace{4pt}}}
        \toprule
        Src \textbackslash\ Tgt & En & Es \\
        \midrule
        En & 67.6 & 52.0 \\
        Es & 71.4 & 56.4 \\
        \bottomrule
    \end{tabular}
    \caption{\label{tab:openassistant-bestofk-human}
    Cross-lingual alignment results using \textbf{\bok} with $n=64$, for the \textbf{dialog generation} task, measured in win rate (\%) against the target-language SFT model as judged by \textbf{human evaluators}~(\autoref{fig:humans}).
    }
\end{table}

\begin{table}[t!]
    \centering
    \begin{tabular}{@{\hspace{4pt}}c@{\hspace{4pt}}|@{\hspace{6pt}}c@{\hspace{6pt}}c@{\hspace{6pt}}c@{\hspace{6pt}}c@{\hspace{6pt}}c@{\hspace{6pt}}c@{\hspace{4pt}}}
        \toprule
        Src \textbackslash\ Tgt & De & En & Es & Ru & Tr & Vi \\
        \midrule
        De & -- & 50.0 & 61.9 & 66.1 & 66.1 & 54.6 \\
        En & 47.9 & -- & 63.3 & 64.9 & 64.5 & 53.1 \\
        Es & 50.6 & 52.9 & -- & 64.1 & 64.5 & 59.0 \\
        Ru & 47.4 & 51.2 & 60.3 & -- & 63.3 & 57.7 \\
        Tr & 50.6 & 52.5 & 61.8 & 65.6 & -- & 50.8 \\
        Vi & 42.0 & 50.8 & 59.1 & 64.4 & 63.6 & -- \\
        \bottomrule
    \end{tabular}
    \caption{\label{tab:summarization-bestofk-translatedrm}
    Alignment quality using RM trained by translating the source language data into the target language using \bok with $n=64$, for the summarization task, measured in win rate (\%) against the target-language SFT model as judged by PaLM-2-L~(\S\ref{sec:transferability-as-alignment-signal}).
    }
\end{table}

\begin{table}[t!]
    \centering
    \begin{tabular}{@{\hspace{4pt}}c@{\hspace{4pt}}|@{\hspace{6pt}}c@{\hspace{6pt}}c@{\hspace{6pt}}c@{\hspace{6pt}}c@{\hspace{6pt}}c@{\hspace{6pt}}c@{\hspace{4pt}}}
        \toprule
        Src \textbackslash\ Tgt & De & En & Es & Ru & Tr & Vi \\
        \midrule
        De & 71.0 & 64.8 & 68.0 & 67.9 & 67.5 & 67.7 \\
        En & 62.2 & 67.4 & 67.9 & 66.3 & 66.5 & 70.8 \\
        Es & 67.4 & 62.7 & 72.3 & 69.7 & 71.4 & 65.2 \\
        Ru & 66.5 & 61.3 & 65.4 & 65.7 & 66.5 & 63.6 \\
        Tr & 66.8 & 64.6 & 68.5 & 69.1 & 73.2 & 68.7 \\
        Vi & 63.0 & 66.7 & 68.6 & 66.5 & 67.8 & 71.3 \\
        \midrule
        Majority & 52.9 & 59.5 & 63.1 & 55.1 & 56.2 & 67.9 \\
        Length & 56.6 & 59.5 & 63.1 & 55.1 & 55.2 & 67.9 \\
        \bottomrule
    \end{tabular}
    \caption{\label{tab:summarization-task-data}
    RM generalizability within the reward modeling task evaluated by accuracy (\%) on in-task validation data for the summarization task, on the six Seahorse languages, as well as the majority baseline and the length baseline (\S\ref{sec:transferability-task-data})~(\autoref{fig:task-data}).
    }
\end{table}

\begin{table}[t!]
    \centering
    \begin{tabular}{@{\hspace{4pt}}c@{\hspace{4pt}}|@{\hspace{6pt}}c@{\hspace{6pt}}c@{\hspace{6pt}}c@{\hspace{4pt}}}
        \toprule
        Src \textbackslash\ Tgt & En & Es & Ru \\
        \midrule
        En & 68.4 & 68.4 & 76.3 \\
        Es & 65.4 & 67.8 & 77.0 \\
        Ru & 56.6 & 63.5 & 64.4 \\
        \midrule
        Length & 66.1 & 68.1 & 71.1 \\
        \bottomrule
    \end{tabular}
    \caption{\label{tab:openassistant-task-data}
    RM generalizability within the reward modeling task evaluated by accuracy (\%) on in-task validation data for the dialog generation task, in three languages, as well as the length baseline (\S\ref{sec:transferability-task-data})~(\autoref{fig:task-data}).
    }
\end{table}

\begin{table}[t!]
    \centering
    \begin{tabular}{@{\hspace{4pt}}c@{\hspace{4pt}}|@{\hspace{6pt}}c@{\hspace{6pt}}c@{\hspace{6pt}}c@{\hspace{6pt}}c@{\hspace{6pt}}c@{\hspace{6pt}}c@{\hspace{4pt}}}
        \toprule
        Src \textbackslash\ Tgt & De & En & Es & Ru & Tr & Vi \\
        \midrule
        De & 0.92 & 0.78 & 0.83 & 0.01 & 0.37 & 1.92 \\
        En & 1.50 & 1.32 & 1.01 & 0.02 & 0.83 & 3.30 \\
        Es & 1.78 & 1.63 & 1.51 & 0.10 & 1.39 & 3.92 \\
        Ru & 0.79 & 0.45 & 0.46 & 0.02 & 0.36 & 1.26 \\
        Tr & 2.20 & 1.91 & 1.83 & 0.15 & 1.34 & 4.28 \\
        Vi & 1.78 & 2.52 & 1.74 & 0.02 & 1.47 & 4.37 \\
        \bottomrule
    \end{tabular}
    \caption{\label{tab:summarization-k}
    KL-divergence of the RL models from the corresponding target-language SFT model for the summarization task~(\autoref{fig:winrate-vs-kl}).
    }
\end{table}

\begin{table}[t!]
    \centering
    \begin{tabular}{@{\hspace{4pt}}c@{\hspace{4pt}}|@{\hspace{5pt}}c@{\hspace{5pt}}c@{\hspace{5pt}}c@{\hspace{5pt}}c@{\hspace{5pt}}c@{\hspace{5pt}}c@{\hspace{4pt}}}
        \toprule
        Lg. & De & En & Es & Ru & Tr & Vi \\
        \midrule
        Mono. & 36.2 & 38.9 & 32.9 & 16.9 & 35.2 & 41.8 \\
        Lg$\to$En & 27.8 & -- & 27.1 & 22.4 & 28.2 & 26.7 \\
        En$\to$Lg & 16.1 & -- & 24.6 & 13.6 & 29.9 & 40.3 \\
        En$\to$Lg$\to$En & 36.5 & -- & 36.1 & 35.4 & 36.5 & 35.8 \\
        Lg$\to$En$\to$Lg & 32.5 & -- & 26.6 & 12.2 & 32.1 & 34.9 \\
        \bottomrule
    \end{tabular}
    \caption{\label{tab:translated-sft-rouge}
    ROUGE-L score when the SFT model is trained using different strategies, either monolingually, translated from a source language, or back-translated into a source language and then back~(\autoref{fig:translated-sft-all}(a)).
    }
\end{table}

\begin{table}[t!]
    \centering
    \begin{tabular}{@{\hspace{4pt}}c@{\hspace{4pt}}|@{\hspace{6pt}}c@{\hspace{6pt}}c@{\hspace{6pt}}c@{\hspace{6pt}}c@{\hspace{6pt}}c@{\hspace{4pt}}}
        \toprule
        Lg. & De & Es & Ru & Tr & Vi \\
        \midrule
        \multicolumn{6}{l}{Target-language SFT; RM transfer only} \\
        \midrule
        Lg$\to$En & 50.8 & 51.2 & 46.5 & 52.9 & 48.2 \\
        En$\to$Lg & 56.4 & 66.1 & 70.7 & 67.2 & 63.1 \\
        \midrule
        \multicolumn{6}{l}{(Back-)Translated SFT} \\
        \midrule
        Lg$\to$En & 36.6 & 26.6 & 29.8 & 37.5 & 31.8 \\
        En$\to$Lg & 14.4 & 43.5 & 43.9 & 47.1 & 41.6 \\
        En$\to$Lg$\to$En & 42.7 & 43.2 & 40.1 & 41.4 & 37.1 \\
        Lg$\to$En$\to$Lg & 45.3 & 54.0 & 60.1 & 61.7 & 51.1 \\
        \bottomrule
    \end{tabular}
    \caption{\label{tab:translated-sft}
    Alignment performance using \bok, measured in the win rate against the monolingual target language SFT model as judged by PaLM-2-L, when the SFT model is trained using different strategies. The first section uses a SFT model that is trained on target-language datasets (same as \autoref{tab:summarization-bestofk}), while the second uses translated or back-translated SFT data~(\autoref{fig:translated-sft-all}(b)).
    }
\end{table}

\begin{table}[t!]
    \centering
    \begin{tabular}{@{\hspace{4pt}}c@{\hspace{4pt}}|@{\hspace{6pt}}c@{\hspace{6pt}}c@{\hspace{6pt}}c@{\hspace{6pt}}c@{\hspace{6pt}}c@{\hspace{4pt}}}
        \toprule
        Lg. & De & Es & Ru & Tr & Vi \\
        \midrule
        \multicolumn{6}{l}{Target-language SFT; RM transfer only} \\
        \midrule
        Lg$\to$En & 38.3 & 32.4 & 38.6 & 32.9 & 29.2 \\
        En$\to$Lg & 62.8 & 59.4 & 53.7 & 47.4 & 66.4 \\
        \midrule
        \multicolumn{6}{l}{(Back-)Translated SFT} \\
        \midrule
        Lg$\to$En & 40.5 & 29.1 & 33.2 & 26.0 & 19.4 \\
        En$\to$Lg & 45.7 & 50.3 & 60.3 & 37.1 & 67.6 \\
        En$\to$Lg$\to$En & 31.4 & 33.9 & 34.0 & 40.8 & 31.7 \\
        Lg$\to$En$\to$Lg & 40.3 & 31.2 & 40.1 & 45.9 & 61.4 \\
        \bottomrule
    \end{tabular}
    \caption{\label{tab:translated-sft-wikilingua}
    Alignment performance using \bok, measured in the win rate against the monolingual target language SFT model as judged by PaLM-2-L, when the SFT model is trained using different strategies. The first section uses a SFT model that is trained on target-language datasets, while the second uses translated or back-translated SFT data. Here, we only consider the WikiLingua dataset for both SFT and RM~(\autoref{fig:translated-sft-all}(c)).
    }
\end{table}

\begin{table}[t!]
    \centering
    \begin{tabular}{@{\hspace{4pt}}c@{\hspace{4pt}}|@{\hspace{6pt}}c@{\hspace{6pt}}c@{\hspace{6pt}}c@{\hspace{6pt}}c@{\hspace{6pt}}c@{\hspace{4pt}}}
        \toprule
        Lg. & De & Es & Ru & Tr & Vi \\
        \midrule
        \multicolumn{6}{l}{Target-language SFT; RM transfer only} \\
        \midrule
        Lg$\to$En & 61.0 & 56.1 & 55.2 & 54.6 & 61.8 \\
        En$\to$Lg & 55.9 & 58.5 & 52.6 & 54.8 & 56.6 \\
        \midrule
        \multicolumn{6}{l}{(Back-)Translated SFT} \\
        \midrule
        Lg$\to$En & 60.2 & 37.5 & 22.7 & 54.9 & 19.2 \\
        En$\to$Lg & 28.8 & 57.0 & 56.5 & 59.6 & 51.9 \\
        En$\to$Lg$\to$En & 47.5 & 46.7 & 42.1 & 42.4 & 48.3 \\
        Lg$\to$En$\to$Lg & 44.7 & 45.1 & 46.6 & 49.5 & 30.7 \\
        \bottomrule
    \end{tabular}
    \caption{\label{tab:translated-sft-rl}
    Alignment performance using RL, measured in the win rate against the monolingual target language SFT model as judged by PaLM-2-L, when the SFT model is trained using different strategies. The first section uses a SFT model that is trained on target-language datasets, while the second uses translated or back-translated SFT data~(\autoref{fig:translated-sft-all}(d)).
    }
\end{table}

\begin{table}[t!]
    \centering
    \begin{tabular}{@{\hspace{4pt}}c@{\hspace{4pt}}|@{\hspace{6pt}}c@{\hspace{6pt}}c@{\hspace{6pt}}c@{\hspace{6pt}}c@{\hspace{6pt}}c@{\hspace{6pt}}c@{\hspace{4pt}}}
        \toprule
        Src \textbackslash\ Tgt & De & En & Es & Ru & Tr & Vi \\
        \midrule
        De & 52.3 & 50.8 & 63.0 & 66.7 & 63.0 & 60.4 \\
        En & 56.4 & 55.5 & 66.1 & 70.7 & 67.2 & 63.1 \\
        De + En & 56.6 & 55.7 & 66.6 & 70.6 & 66.7 & 64.1 \\
        \bottomrule
    \end{tabular}
    \caption{\label{tab:bilingual-rm}
    Alignment performance using \bok, measured in the win rate against the monolingual target language SFT model as judged by PaLM-2-L, when using either a monolingual RM (same as \autoref{tab:summarization-bestofk}) or a bilingual RM~(\autoref{fig:bilingual-rm}).
    }
\end{table}